\documentclass{article}

\usepackage{PRIMEarxiv}

\usepackage[utf8]{inputenc} 
\usepackage[T1]{fontenc}    
\usepackage{hyperref}       
\usepackage{url}            
\usepackage{booktabs}       
\usepackage{amsfonts}       
\usepackage{nicefrac}       
\usepackage{microtype}      
\usepackage{lipsum}
\usepackage{fancyhdr}       
\usepackage{graphicx}       
\usepackage{caption}
\usepackage{subcaption}
\usepackage{xcolor}
\usepackage{rotating}
\usepackage[normalem]{ulem}
\usepackage{soul}
\usepackage{amsmath}

\graphicspath{{media/}}     

\title{Do machines fail like humans? A human-centred out-of-distribution spectrum for mapping error alignment
}

\author{
  Binxia Xu \\
  School of Data Science \\
  Fudan University \\
  Shanghai \\
  \texttt{binxia\_xu@fudan.edu.cn} \\
  \And
  Xiaoliang Luo \\
  Independent Researcher\\
 \phantom{Department of Experimental Psychology UCL} \\
  London \\
  \texttt{xiao.luo.17@alumni.ucl.ac.uk} \\
  \AND
  Luke Dickens$^\dagger$ \\
  Department of Information Studies \\
  University College London \\
  London \\
  \texttt{l.dickens@ucl.ac.uk} \\
  \And
  Robert M. Mok$^\dagger$ \\
  Center for Information and Neural Networks  \\
  National Institute of Information and \\ Communications Technology\\
   University of Osaka\\
   Osaka \\
  \texttt{rob.mok@nict.go.jp} \\
}

\begin{document}
\maketitle
\renewcommand{\thefootnote}{\fnsymbol{footnote}}
\footnotetext[2]{Equal senior authorship contribution.}

\begin{abstract}

Determining whether AI systems process information similarly to humans is central to cognitive science and trustworthy AI. While modern AI models can match human accuracy on standard tasks, such parity does not guarantee that their underlying decision-making strategies resemble those of humans. Assessing performance using error alignment metrics to compare how humans and models fail, and how this changes for distorted, or otherwise more challenging, stimuli, provides a viable pathway toward a finer characterization of model-human alignment. However, existing out-of-distribution (OOD) analyses for challenging stimuli are limited due to methodological choices: they define OOD shift relative to model training data or use arbitrary distortion-specific parameters with little correspondence to human perception, hindering principled comparisons.
We propose a human-centred framework that redefines the degree of OOD as a spectrum of human perceptual difficulty. By quantifying how much a collection of stimuli deviates from an undistorted reference set based on human accuracy, we construct an OOD spectrum and identify four distinct regimes of perceptual challenge. This approach enables principled model-human comparisons at calibrated difficulty levels. We apply this framework to object recognition and reveal unique, regime-dependent model-human alignment rankings and profiles across deep learning architectures.  Vision-language models are most consistently human aligned across near- and far-OOD conditions, but convolutional neural networks (CNNs) are more aligned than vision transformers (ViTs) for near-OOD and ViTs are more aligned than CNNs for far-OOD. Our work demonstrates the critical importance of accounting for cross-condition differences, such as perceptual difficulty, for a principled assessment of model-human alignment.
 \end{abstract}


\section{Introduction}

Evaluating whether AI systems process information like humans is a core challenge for cognitive and computer science. Neural networks can serve as computational abstractions to test hypotheses of human processing in cognitive science~\cite{kriegeskorte2015deep,cichy2019deep,pulvermuller2021biological,rajalingham2018large,sucholutsky2023getting}, while in computer science, such alignment improves robustness~\cite{peterson2019human,muttenthaler2025aligning}, interpretability~\cite{fel2022harmonizing,boyd2024increasing}, and trustworthiness~\cite{linsley2023adversarial,hoak2025alignment}. 
Existing AI models achieve or exceed human accuracy on clear inputs. However, human-level accuracy does not entail model-human alignment, as systems may make different errors reflecting completely different processing mechanisms. Thus, going beyond accuracy is crucial~\cite{borji2014human,rajalingham2018large}. Metrics such as error consistency (EC)\cite{geirhos-2020-beyond} and misclassification agreement (MA)\cite{xu2025measuring} address this by measuring error patterns between systems. Yet comparing errors on easy tasks is insufficient -- alignment may appear high on clear stimuli simply due to the infrequency of errors. Challenging conditions, where error patterns can become more distinct, are essential for rigorously assessing model-human alignment.

In previous alignment analyses of visual perception, degraded inputs have revealed inductive biases and information-processing strategies~\cite{geirhos2018imagenet,baker2018deep,conwell2024large}. Research comparing human and model behavior under 'out-of-distribution' (OOD) samples~\cite{kheradpisheh2016humans,dodge2017study,geirhos2018generalisation,hendrycks2019robustness} suggests that distortions to features like texture or global shape differentially impair systems based on the cues they privilege. However, to assess model-human alignment appropriately, performance must be compared on a common and meaningful scale. We identify four fundamental problems for model-human comparison under distorted conditions and provide a human-centred behavioral deviation framework to solve these issues.

\begin{figure*}[t!]
\centering
\includegraphics[width=1\linewidth]{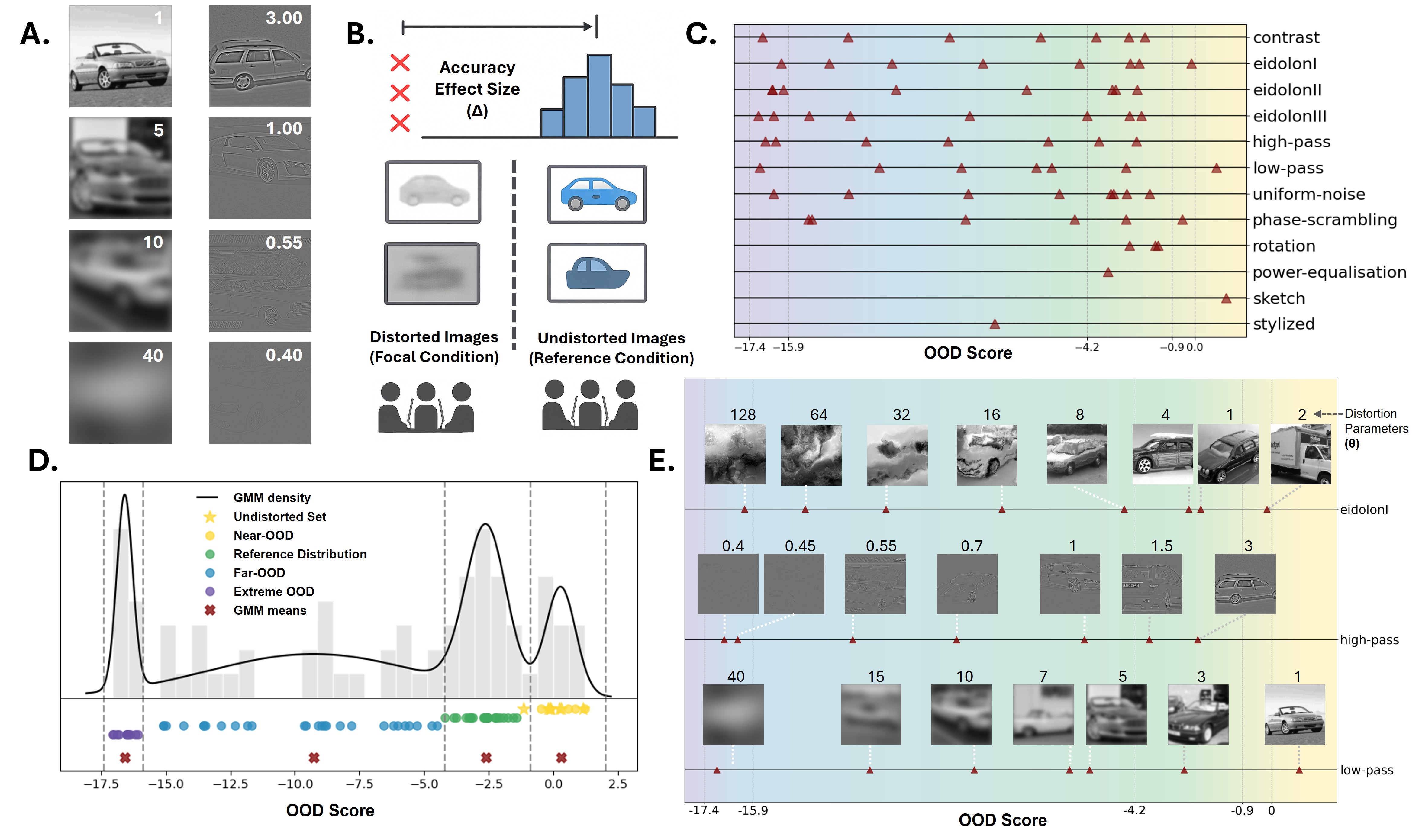}
\caption{A human-centered OOD spectrum. (A) Example images from two distortion types (left column: low-pass filtering; right column: high-pass filtering) from the \textit{modelvshuman} datase~\cite{geirhos2021partial} at different parameter levels. Images within the same row share the same parameter rank within each distortion type. (B) Illustration of the process for measuring the degree of human performance deviation on a specific distortion condition relative to undistorted images. (C)  OOD scores for all levels of corruption for each distortion type in \textit{modelvshuman} dataset. The x-axis represents the OOD score, indicating the degree of deviation from the performance baseline on undistorted images, with greater negative values corresponding to larger distributional shifts.  (D) Four-component Gaussian Mixture Model (GMM) fitted to OOD scores across all distortion conditions, defining the four OOD groups. Color reflects different OOD regimes. Boundaries illustrated by dashed lines. (E) OOD spectrum with example images from selected distortions.}
\label{fig:spectrum}
\end{figure*}

First, \textbf{there is no clear meaning to OOD stimuli in humans}. In machine learning, OOD is defined by deviation from training data statistics~\cite{farquhar2022out}. Recent work on model-human alignment is guided by this definition~\cite{hendrycks2016baseline,hendrycks2019robustness,taori2020measuring}, but humans do not have a finite, controlled ``training distribution'', and instead acquire knowledge through a lifetime of varied, naturally encountered experience. This asymmetry raises fundamental questions:  How should we define OOD for human perception? For instance, which images in
Figure~\ref{fig:spectrum}A are OOD for humans?  

Second, \textbf{there is no unified principled measure of degree of distortion in OOD stimuli}.
Consider the distortions in Figure~\ref{fig:spectrum}A, how do we compare the degree of distortion between images in the first row and those in the second? When investigating model-human alignment in distorted images, some studies aggregate performance across severity levels ~\cite{hendrycks2019robustness,geirhos2021partial,xu2025measuring}, but such an approach ignores how perceptual difficulty varies across a distortion type. Others qualitatively select distortion levels that appear roughly comparable ~\cite{dodge2016understanding,dodge2017study,wichmann2017methods,malik2023extreme}, but this introduces subjective judgement. Hendrycks et al. ~\cite{hendrycks2019robustness} identified the problem with grouping distortion conditions with different difficulties and address this by normalising to a reference model, but this substitutes one arbitrary reference for another where the resulting scale reflects the idiosyncrasies of that architecture rather than any principled notion about image degradation.

Third, \textbf{not all degradation conditions are equally meaningful} (as raised in \cite{geirhos2021partial}). Consider an extreme example where an image with a low-pass parameter of 40 appears unrecognizable to humans (Figure~\ref{fig:spectrum}A) -- would such a stimulus be appropriate for assessing model-human alignment? If not, how should we determine the appropriate stimulus set? 

Lastly, \textbf{raw model-human alignment values lack meaning without a baseline}. Previous work treats alignment values directly as evidence of alignment~\cite{geirhos2021partial}. However, this interpretation is incomplete without considering a baseline of such human–human alignment values. When humans show zero agreement on a stimulus set, we cannot expect high model-human alignment.

To address these challenges, we propose a \textbf{human-centred behavioral deviation framework} that provides a natural benchmark for fairly comparing model-human alignment across distortion types on a common scale. Rather than using distortion-specific parameters (Figure~\ref{fig:spectrum}A), we place them on a shared scale anchored in human perceptual difficulty (Figure~\ref{fig:spectrum}C). 
We operationalise this \textbf{OOD score} using the effect size Glass's $\Delta$~\cite{hedges1981distribution} applied to accuracy logits -- deviation of human performance from undistorted stimuli, which we term the reference distribution.
With this, we define OOD for humans as deviation from the reference condition, providing a quantitative scale we call the perceptual OOD spectrum. Defining OOD in terms of human perceptual difficulty provides a principled way to control task difficulty across conditions, enabling systematic investigation of visual processing strategies that would be confounded by arbitrary distortion parameters~\cite{kayargadde1996perceptual}.

Our analysis is conducted on the \textit{modelvshuman} dataset~\cite{geirhos2021partial}, which contains human core object recognition performance on systematically distorted images. To preview the results, our approach reveals that a perceptual OOD spectrum contains distinct regions that reflect different regimes of human information processing. We evaluate models within near- and far-OOD regions, showing distinct model-human alignment patterns across regimes: Vision-Language Models (VLMs)~\cite{radford2021learning} show the most consistent alignment across regimes, whereas Vision Transformer models (ViTs)~\cite{dosovitskiy2020image} achieve higher alignment than Convolutional Neural Networks (CNNs)~\cite{simonyan2014very} in near-OOD but the reverse pattern for far-OOD. 
However, despite strong alignment in some models, human–human alignment remains a ceiling that no model approaches, indicating that the visual features humans rely on overlap more with each other than with those exploited by models. 
In sum, this work introduces a human-centred framework to appropriately compare human and model error profiles, and offer an empirical characterization of how different architectures align with human perception across regimes.

\section{Results}

\subsection{Constructing an Out-of-Distribution (OOD) Spectrum Based on Human Perceptual Difficulty}

\subsubsection{Human Behavior Shifts Under Distortions}

To test whether distortions produce significant shifts in human recognition behavior, we compared the distribution of human accuracies under distortion for each condition (one condition refers to one level of one distortion type, four participants per condition) with the undistorted baseline (28 values in total) (Figure~\ref{fig:spectrum}.A). We found that distorted images induced substantial shifts in behavior: only 7 out of 65 distortion conditions were not  significantly different from the undistorted condition (Mann–Whitney U test: $p<0.01$, False Discovery Rate corrected; see
Table~S\ref{tab:mw_test} in Appendix A for all results), and accuracy was significantly above chance in 54 of 65 conditions ($P$=1/16 categories; one-tailed binomial tests, FDR corrected; see Table~S\ref{tab:binomial_test} for all results).

\subsubsection{Building the OOD Spectrum through Quantifying Human Performance Deviation}

To build the perceptual OOD spectrum, we construct a reference distribution that quantifies how much each distorted condition deviates from the undistorted baseline (i.e., the effect size), providing a continuous measure of distortion severity grounded in human performance.
Due to non-normality in the accuracy scores (Shapiro–Wilk test, $p < 0.05$), we applied a logit transformation, after which all tests yielded $p > 0.05$, indicating no rejection of the null hypothesis of normality (see Appendix A, Table S\ref{tab:normality_tests}).

To characterize the distribution of human accuracy under distortion, we define an \textbf{OOD score} based on Glass’s $\Delta$ as the measure for behavioral deviation:
\begin{equation}
\Delta = \frac{\bar{l_d} - \bar{l_{ud}}}{s_{ud}},
\label{eq:glass-delta}
\end{equation}
where $\bar{l_d}$ and $\bar{l_{ud}}$ are the mean logit-transformed accuracies on the distorted set and undistorted set respectively, and $s_{ud}$ is the standard deviation for the logit-transformed accuracy on the undistorted images. 
Glass's $\Delta$ is appropriate because it standardises the mean difference between performance on distorted and undistorted images using only the reference condition's variance, ignoring the variance of each distorted condition. This ensures comparability across conditions by measuring all deviations on a common scale defined by the reference -- the baseline condition -- and avoids underestimating behavioral differences for distortion conditions with high variance.

%

\subsubsection{The Perceptual Distortion OOD Spectrum}

\begin{figure*}[tbhp]
\centering
\includegraphics[width=0.9\linewidth]{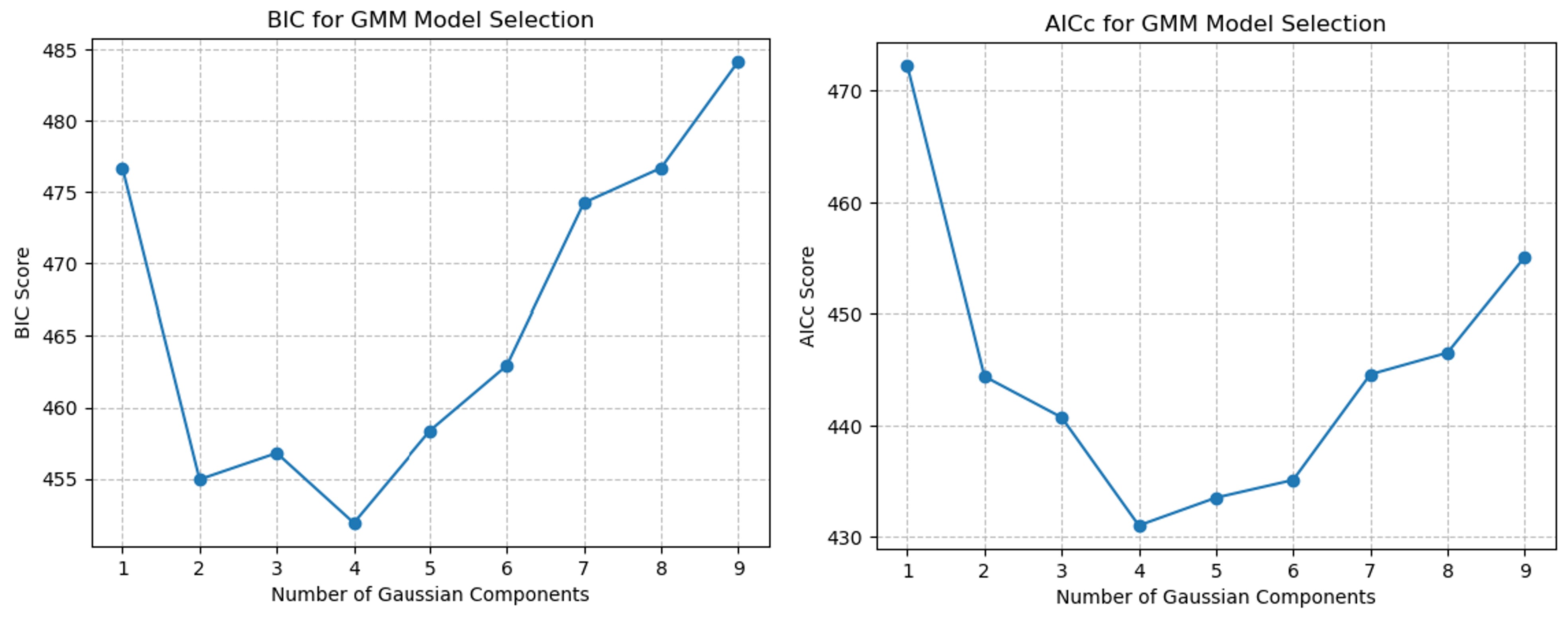}
\caption{Model selection using the Bayesian Information Criterion (BIC) and corrected Akaike Information Criterion (AICc). Both criteria indicate four components as the optimal model complexity, with BIC and AICc reaching their minimum values at this point.}
\label{fig:bic}
\end{figure*}

To meaningfully assess model robustness in relation to human perception, we construct a human-centred OOD spectrum based on behavioral deviation across distortion types and levels, to replace the arbitrary distortion parameter scales.
To illustrate the range of behavioral deviation, we present three example distortion types (Figure~\ref{fig:spectrum}E).
The behavioral deviation effect sizes for all distortion conditions are presented in Figure~\ref{fig:spectrum}C.
To explore error profiles and investigate the distinct inductive biases by different distortion types, we used a data-driven approach to group conditions that produce comparable degrees of behavioral deviation. We fitted a Gaussian Mixture Model (GMM) to the distribution of OOD scores across all distortion types and levels (including the undistorted sets), which consistently gave a four-component solution (using Bayesian Information Criterion (BIC) or Akaike Information Criterion (AICc); see Figure S~\ref{fig:bic}).

We interpret these groups as qualitatively different regimes of perceptual difficulty (Figure~\ref{fig:spectrum}D). 
We label these groups as \textbf{reference}, \textbf{near-OOD}, \textbf{far-OOD}, and \textbf{extreme-OOD}, ordered by group means. 
Notably, the conditions that fall within the extreme-OOD level correspond precisely to those yielding chance level human performance, suggesting little or no recognisable information and are potentially of less interest than the other OOD regimes.

We interpret these groups as qualitatively different regimes of perceptual difficulty (Figure~\ref{fig:spectrum}.D). 
We label these groups as reference, near-OOD, far-OOD, and extreme-OOD, ordered by group means. 
Notably, the conditions that fall within the extreme-OOD level correspond precisely to those yielding chance level human performance, suggesting little or no recognisable information and are potentially of less interest than the other OOD regimes.

\subsection{Alignment in Human Perceptual Failures}

Although the spectrum is constructed based on accuracy, accuracy alone only provides one aspect of behavior -- it indicates the degree of recognition failure, but not \emph{how} it fails. Two distortions can yield similar accuracy effect sizes yet produce very different behavioral patterns; participants may show similar accuracy with highly similar error profiles or with entirely different errors. 
To capture this dimension, we examine human–human error alignment, quantifying how consistently participants misclassify the same images or choose the same incorrect labels. Specifically, we use two complementary metrics: Error Consistency (EC)~\cite{geirhos-2020-beyond}, which measures the overlap of misclassified samples between pairs of participants, and Misclassification Agreement (MA)~\cite{xu2025measuring}, which measures how often participants predict the same class on samples when both are incorrect.

\subsubsection{ Error alignment across OOD regimes are markedly different}
\begin{figure*}[tbhp]
\centering
\includegraphics[width=0.95\linewidth]{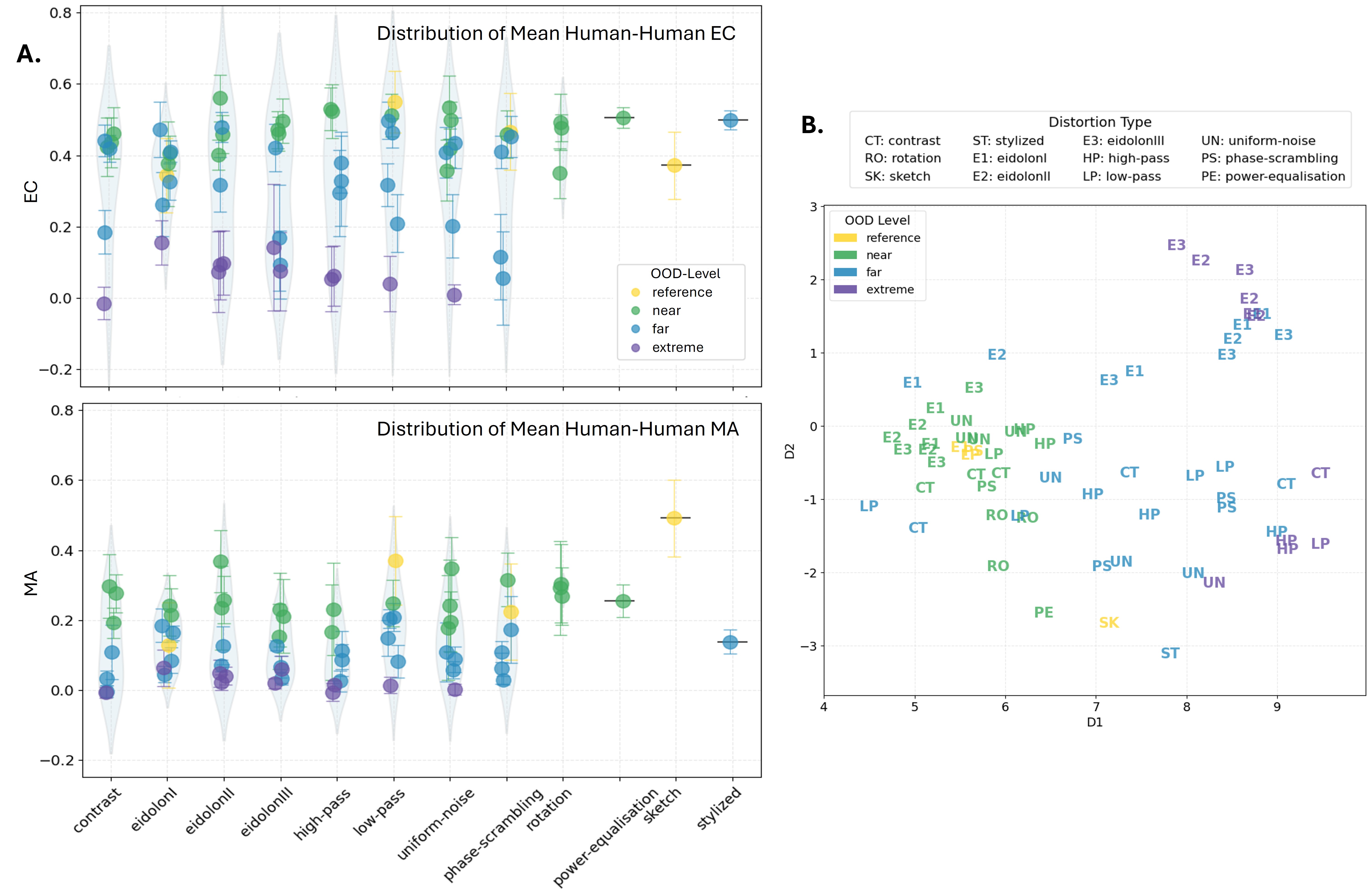}
\caption{ Human-human error alignment A. Dot plots for Error Consistency (EC; top), and Misclassification Agreement (MA; bottom) across distortion types and OOD levels. Each dot represents the mean error alignment value between each pair of participants for a specific level of a distortion, coloured by its OOD category, with vertical bars indicating ±1 standard deviation.
Transparent violin plots show the distributions of the alignment scores across distortion levels within distortion type.
B. t-SNE plot illustrating similarity of human-human error patterns across distortion domains based on the human-human class-level error divergence (CLED) matrix. Each point represents a distortion condition, labeled by distortion type (e.g., CT: contrast, HP: high-pass).  Colors indicate OOD regime they belong to. Spatial proximity reflects similarity in human error patterns across conditions. }
\label{fig:hh_error_align}
\end{figure*}
We assess human-human error alignment across the four OOD regimes (Figure~\ref{fig:hh_error_align}A). 
We find that extreme-OOD distortions consistently yielded much lower EC values compared to near- and far-OOD conditions, highlighting a pronounced breakdown in shared perceptual strategies under severe distortion (Figure~\ref{fig:hh_error_align}A, top). 
As noted, all conditions in the extreme-OOD region are consistent with chance level accuracies. Distortions also differed in the rate at which EC declines across OOD levels. For instance, \textit{uniform noise} and \textit{phase scrambling} show a sharp drop of EC in the far-OOD regime, whereas distortions such as \textit{high-pass} decrease more gradually.
Similar to EC, MA declines with increasing perceptual difficulty, with extreme-OOD levels consistently producing the lowest scores (Figure~\ref{fig:hh_error_align}A, bottom). 
In contrast, the difference between far-OOD and extreme-OOD for MA is less pronounced than for EC in some distortion types such as \textit{eidolon} and \textit{high-pass}. The ordering of alignment values across OOD regimes from near to extreme mirrors EC closely, reinforcing the robustness of our perceptual difficulty–based spectrum.
The violin plots complement these findings by showing that MA distributions are generally more concentrated than EC distributions and are typically centred at lower values. 

\subsubsection{
Human error profiles are primarily structured by perceptual difficulty (OOD score)}



\begin{table}[htbp]
\centering
\begin{tabular}{c|ccc}
\toprule
 & Cohens'$D$  & \multicolumn{2}{c}{Permutation Test} \\
\cmidrule(lr){3-4}
 &  & $p$-value & Effect Size \\
\midrule
Distortion Type & -0.161& 0.025 & -2.331 \\
\midrule
OOD-Level & -0.599 & 0.000 & -12.555 \\
\bottomrule
\end{tabular}
\vspace{3mm}
\caption{Comparison of distortion type and OOD level effects on class-level error divergence (CLED). Reported are Cohen’s $D$ and permutation test results. Negative $D$ indicates lower within-group CLED (greater error similarity). OOD level shows a markedly stronger effect than distortion type in both measures.}
\label{tab:cled_quant}
\end{table}

To further assess whether different severity levels within the same distortion type should be treated as distinct OOD conditions, we compare human error profiles across each OOD level using CLED~\cite{xu2025measuring}. 
Similar to MA, CLED measures the similarity in error patterns between two sets of behavioral responses, but with key differences. As it is a disimilarity measure, larger values suggest greater difference. More importantly, CLED compares responses at the class level using modified confusion matrices, meaning that there is no requirement for the two response sets to contain the same stimuli, allowing us to directly compare how error profiles vary not only within a distortion type but also across different types and OOD levels.

Previous studies have often treated each distortion type (e.g., \textit{uniform noise} or \textit{low-pass filtering}) as a monolithic condition~\cite{geirhos2018generalisation}.  
Nonetheless, distortions that differ in visual form can impose comparable perceptual challenges and elicit similar error profiles. To capture these relationships, we define each domain as a specific distortion type at a given severity level and construct a CLED-based distance matrix across all domains. This analysis allows us to assess the relative contributions of distortion type and OOD severity to the structure of human error behavior.

The similarity of error profiles across human participants is visualized using t-SNE (Figure~\ref{fig:hh_error_align}B). If different severity levels within a distortion type produced similar error profiles, we would expect tight, coherent clusters for each distortion type -- conditions of the same type would group together regardless of severity. However, this is not always the case: lower and higher severity levels within the same distortion type can be far apart, while conditions from different distortion types sometimes appear more closely aligned. Notably, conditions within the near-OOD regime (green labels) cluster tightly together, whereas their counterparts in far- (blue) or extreme-(purple) OOD regimes are typically more dispersed.

We confirm this through quantitative analyses comparing the effect of distortion type and level. 
We estimate the separability of within- versus between-group CLED values for distortion-type-based grouping and OOD-level-based grouping using Cohen’s $D$ and performed permutation tests for statistical inference. The negative values indicate that error patterns of samples within a group (e.g., an OOD level or a distortion type) are more internally consistent than between groups (Table~\ref{tab:cled_quant}). 
Across both measures, OOD level consistently exhibited a much stronger effect than distortion type, reinforcing that perceptual difficulty is a crucial factor influencing human error profiles.

\subsection{Model Error Profiles Across the OOD Spectrum: The Role of Distortion Type, Level, and Model Architecture}

Human behavior provides a foundation for model–human comparison: the degree of human-human alignment establishes an empirical upper bound on achievable model-human alignment. Here, we examine model–human alignment across OOD regimes and which architectures show more human-like error patterns. 

\begin{figure*}[tbhp]
\centering
\includegraphics[width=1\linewidth]{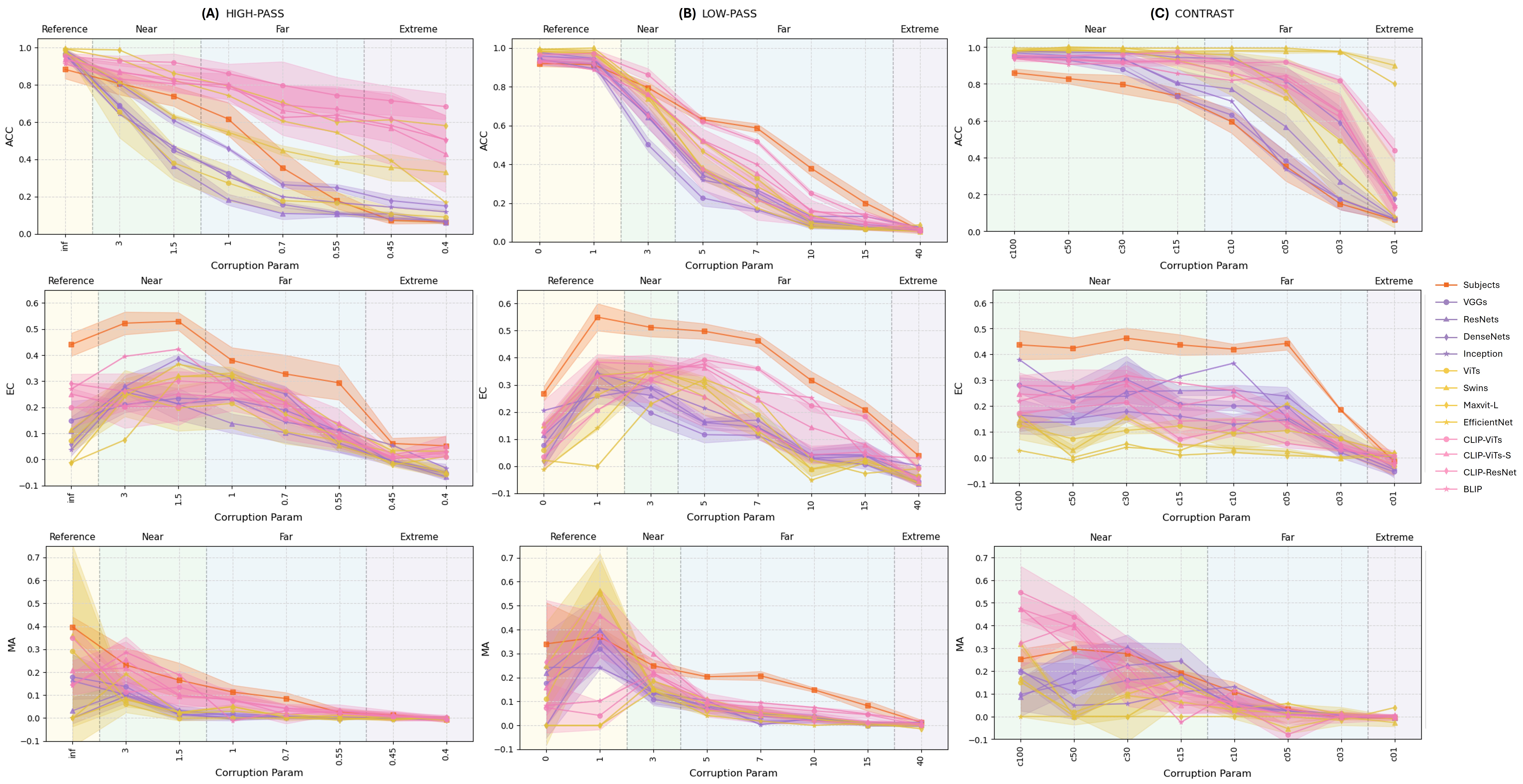}
\caption{Model-human alignment across distortion levels. Accuracy (ACC, top row), Error Consistency (EC, middle row), and Misclassification Agreement (MA, bottom row) for humans and models across three representative distortion types: High-pass (left), Low-pass (middle), and Contrast (right). Each curve shows the mean: for humans, the mean across participant pairs (orange); for models, the mean across model–human pairs within a subfamily. Curve colours denote superfamilies (CNNs: purple, ViTs: yellow, VLMs: pink), while marker shapes distinguish subfamilies within each superfamily. Shaded regions indicate ±1 standard deviation. Curves are plotted against corruption parameters that control the severity, with background shading marking OOD regimes.}
\label{fig:model_curves}
\end{figure*}

\subsubsection{Patterns and variability in model–human alignment across OOD levels} 
To examine how alignment changes over distortion severity and OOD levels, we plot accuracy, EC, and MA patterns across the full range of distortion levels for three distortion types (Figure~\ref{fig:model_curves}, \textit{High-Pass}, \textit{Low-Pass} and \textit{Contrast}; see Figure~S\ref{fig:model_curves_eid} and~S\ref{fig:model_curves_last} for other distortion types). In general, we observe high variability across OOD levels.

Accuracy performance and alignment decline as we move from the reference distribution through near- to far-OOD  -- models broadly track the human trend but differ in where and how they deviate. In the \emph{reference} regime (yellow background), models typically exceed human accuracy, though \textit{Contrast} does not have a condition that falls within this region. For some curves, we see an increase in EC in the near OOD region compared to the reference distribution. It is unclear precisely why this is and further investigation may be required. MA is also less informative here due to there being very few errors.
In the \emph{near-OOD} regime (green background) human–human EC remains higher than model-human EC across distortion types. Importantly, EC slightly increases despite reduced accuracy for \textit{High-Pass} and \textit{Contrast}, indicating that people fail on a shared set of samples as perceptual difficulty increases within this regime. 
MA provides a complementary view: across all distortion types, several models approach human–human MA, meaning that their incorrect predictions resemble human errors. Notably, MA is a stricter criterion than EC, so numerical differences tend to be smaller than in EC.

Entering \emph{far-OOD} regime (blue), under \textit{Low-Pass}, humans often outperform models, whereas  \textit{High-Pass} and \textit{Contrast} largely preserve their near-OOD ordering. 
Human–human EC under \textit{High-Pass} becomes more variable (shaded error bars) yet mostly remains above model-human EC. For \textit{Low-Pass}, human–human EC stays higher, though its margin over model-human EC narrows. 
Of note, EC for humans for \textit{Contrast} is exceptionally high (see level \textit{C05}), reflecting strong agreement about some images even at higher distortion levels. 
MA under \textit{Contrast} is comparatively low, meaning that both humans and models may agree on the correct-incorrect boundary (high EC) but diverge on their mistakes. By contrast, \textit{Low-Pass} produces the largest gap between human–human and model–human MA.

Although alignment degrades with OOD level, the degree of degradation is strongly model-dependent.  VLMs tend to maintain higher MA, CNNs struggle under \textit{Low-Pass}, and ViT-based architectures like Swin and MaxViT show robustness under \textit{High-Pass}. However, these patterns vary across OOD regimes, making it difficult to draw conclusions about alignment across the full spectrum. Therefore, we focus on quantifying regime-specific alignment in the following sections.

\subsubsection{Architectural family effects on human alignment}
\begin{figure*}[tbhp]
\centering
\includegraphics[width=1\linewidth]{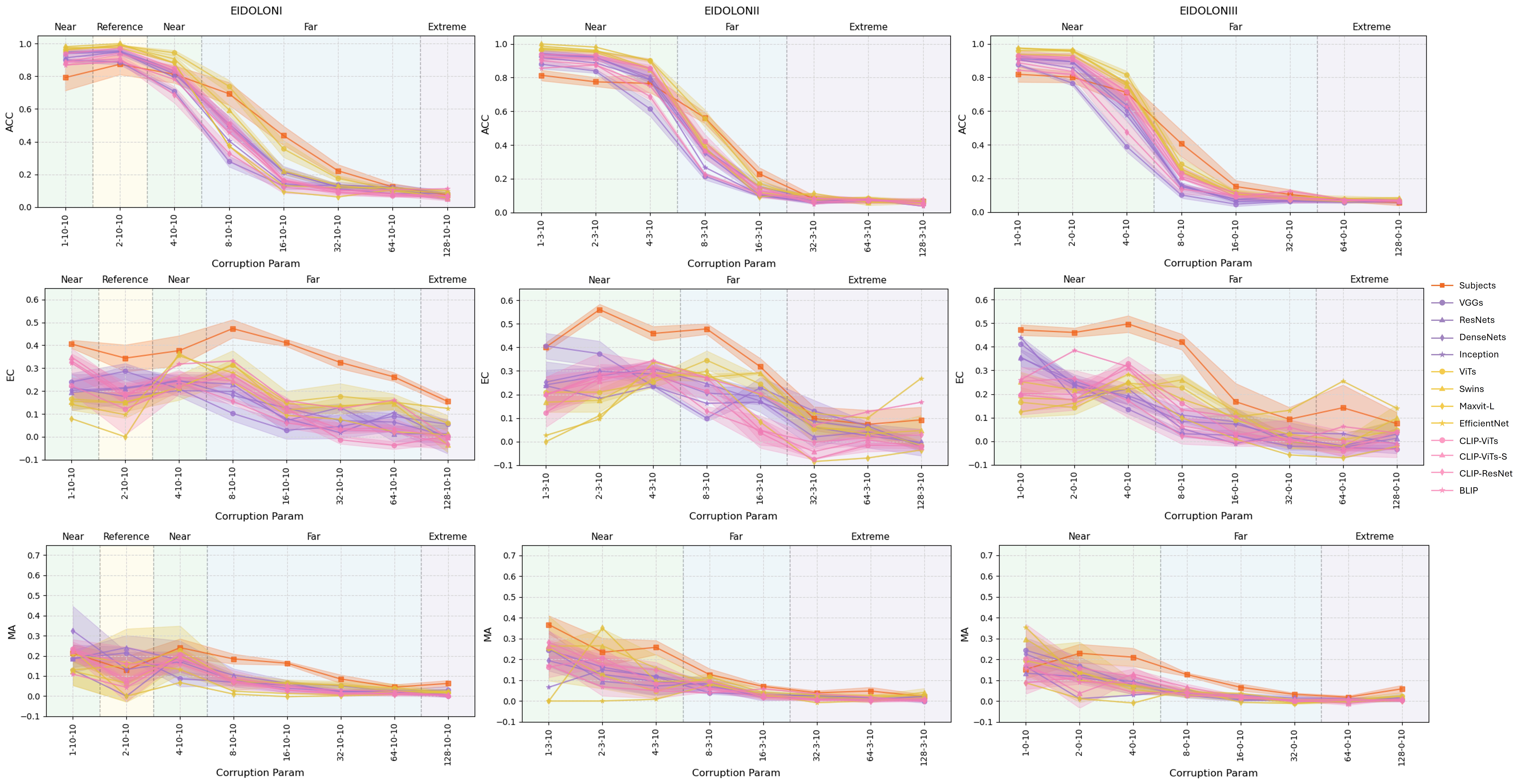}
\caption{Accuracy (ACC, top row), Error Consistency (EC, middle row), and Misclassification Agreement (MA, bottom row) for humans and models across three distortion types: EidolonI (left), EidolonII (middle), EidolonIII (right). }
\label{fig:model_curves_eid}
\end{figure*}

\begin{figure*}[tbhp]
\centering
\includegraphics[width=1\linewidth]{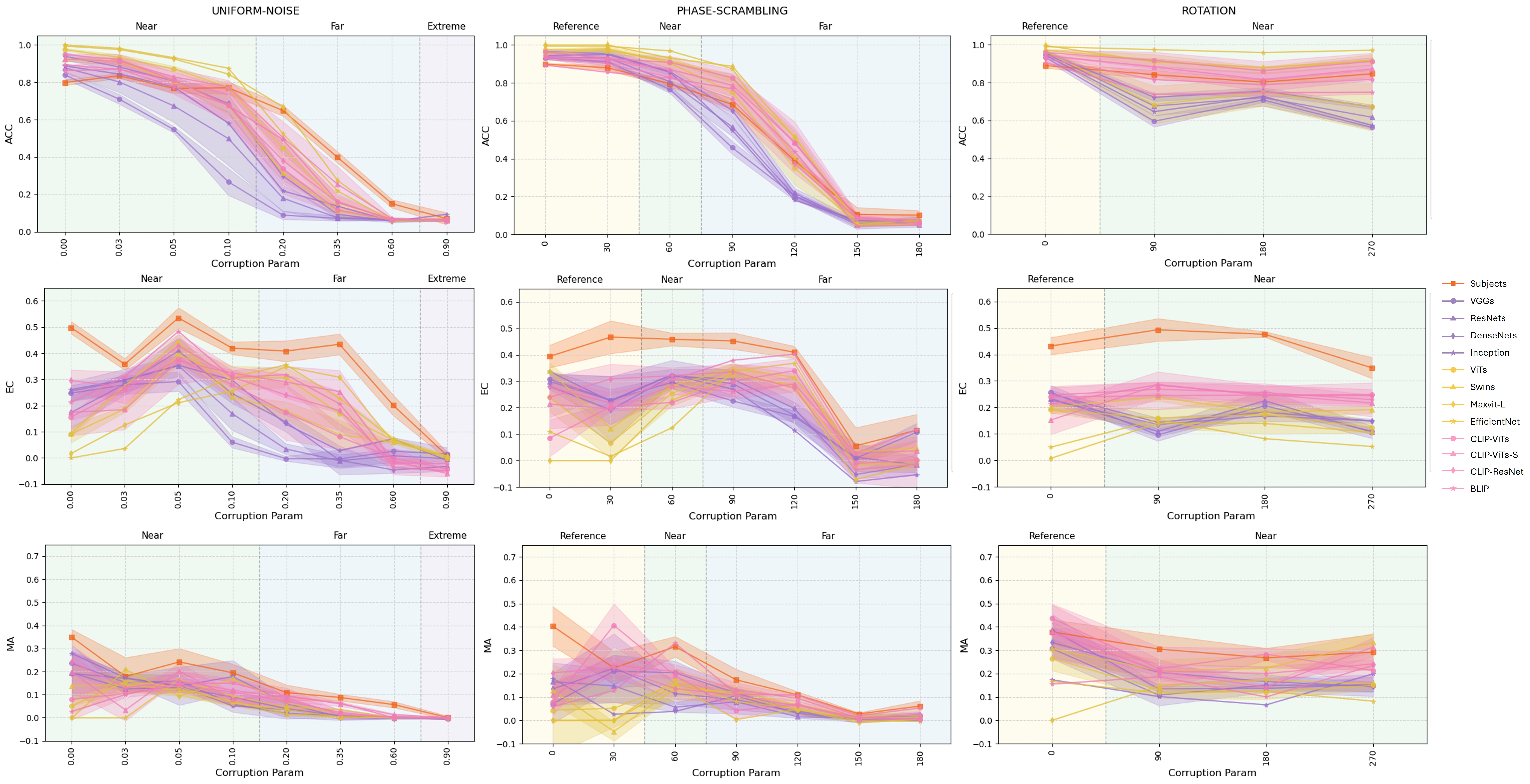}
\caption{Accuracy (ACC, top row), Error Consistency (EC, middle row), and Misclassification Agreement (MA, bottom row) for humans and models across three distortion types: Uniform Noise (left), Phase Scrambling (middle), Rotation (right). }
\label{fig:model_curves_last}
\end{figure*}

Based on qualitative observation of Figure~\ref{fig:model_curves} (see also Figure S~\ref{fig:model_curves_eid}, S~\ref{fig:model_curves_last}), models belonging to the same architectural family tend to exhibit characteristic error-alignment profiles.
This motivates a natural question: are models architecturally closer to one another also more similar in how they align with humans? For example, are the two models from the VGG family more similar in their alignment than they are to other members of the wider CNNs? 
We formalise \textit{two hypotheses}: \textit{H1)}  model-human alignment is more similar between members of the same sub-family than between members of the same super-family; \textit{H2)} model-human alignment is more similar between members of the same super-family than across super-families.
Here, we define two levels of architectural families: sub-family (e.g., VGGs, ResNets) and super-family (e.g., CNNs, ViTs).

We test these hypotheses using permutation tests on model-human alignment profiles. For each model, we concatenate EC and MA values across all distortion types into a single alignment vector and compute pairwise Euclidean distances between models. For \textit{H1}, we compare within-sub-family distances (e.g., VGG-16 vs. VGG-19) to across-sub-family distances within the same super-family (e.g., VGGs vs. ResNets). For \textit{H2}, we compare within-super-family distances to across-super-family distances.
The $p$-values and the effect size are reported in Table~\ref{tab:inter-vs-intra-family-alignment}. 
At the super-family level, under both OOD regimes, this effect is statistically robust ($p < 0.01$) and with strong effect sizes.
At the sub-family level, results depends upon architectures and OOD regimes. 
Generally, far-OOD leads to a stronger effect, indicating greater alignment similarity within a model family. 
CNNs strongly reject the null hypothesis in both near- and far-OOD regimes, showing the strongest effect sizes among the three superfamilies.
By contrast, ViTs do not exhibit statistically reliable sub-family clustering in either regime, suggesting that architectural variations within the ViT family may give rise to more diverse error patterns.
For VLMs, the effect in far-OOD is higher than that in near-OOD.

\begin{table}

\centering
\begin{tabular}{lp{7cm}cccc}
\toprule
Model  &  & \multicolumn{2}{c}{near-OOD} & \multicolumn{2}{c}{far-OOD} \\
Family & Families & p-value & Effect Size & p-value & Effect Size \\
\midrule
All & CNN, ViT, VLM
    & 0.001 & 3.721 & $<$5e-4 & 14.923 \\
CNN & VGG, ResNet, DenseNet, Inception*
    & 0.005 & 3.072 & $<$5e-4 & 4.593 \\
ViT & ViT (l or b), Swin, MaxVit*, EfficientNet*
    &0.336 & 0.424 & 0.057 & 2.037 \\
VLM & CLIP-ResNet, CLIP-ViT, CLIP-Vit (small), BLIP*
    &0.014 & 2.908 & 0.005 & 3.323 \\
\bottomrule
\end{tabular}
\caption{Results of inter- vs intra-family alignment permutation test for near-OOD and far-OOD. Families marked with an asterisk (*) contain one model so do not contribute to sub-family differences. }
\label{tab:inter-vs-intra-family-alignment}
\addtocounter{table}{-1} 

\end{table}

\subsubsection{Profiles of model-human error alignment over distortion types}
\begin{figure*}[tbhp]
\centering
\includegraphics[width=0.9\linewidth]{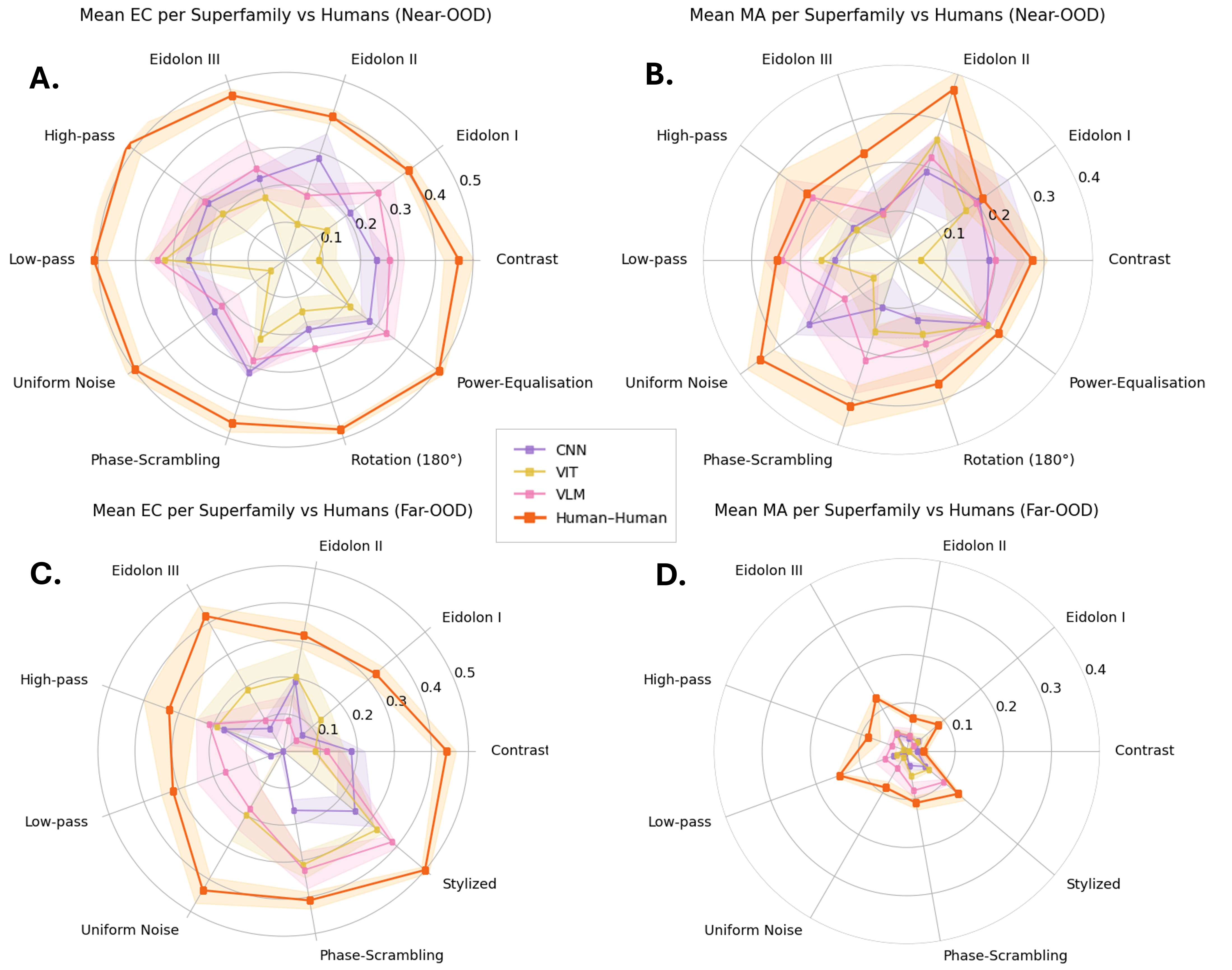}
\caption{Radar plots of mean model–human error alignment for each superfamily across distortion types, separated by OOD regime (near-OOD, EC (A), MA (B); far-OOD, EC (C), MA (D)). Each axis represents a distortion type, and the radial coordinate indicates model–human alignment scores. Each curve shows the mean EC or MA value for a model superfamily, with error bars indicating ±1 standard deviation. Human–human alignment scores are denoted in orange.} 
\label{fig:radar}
\end{figure*}

\begin{figure*}[tbhp]
\centering  
\includegraphics[width=0.9\linewidth]{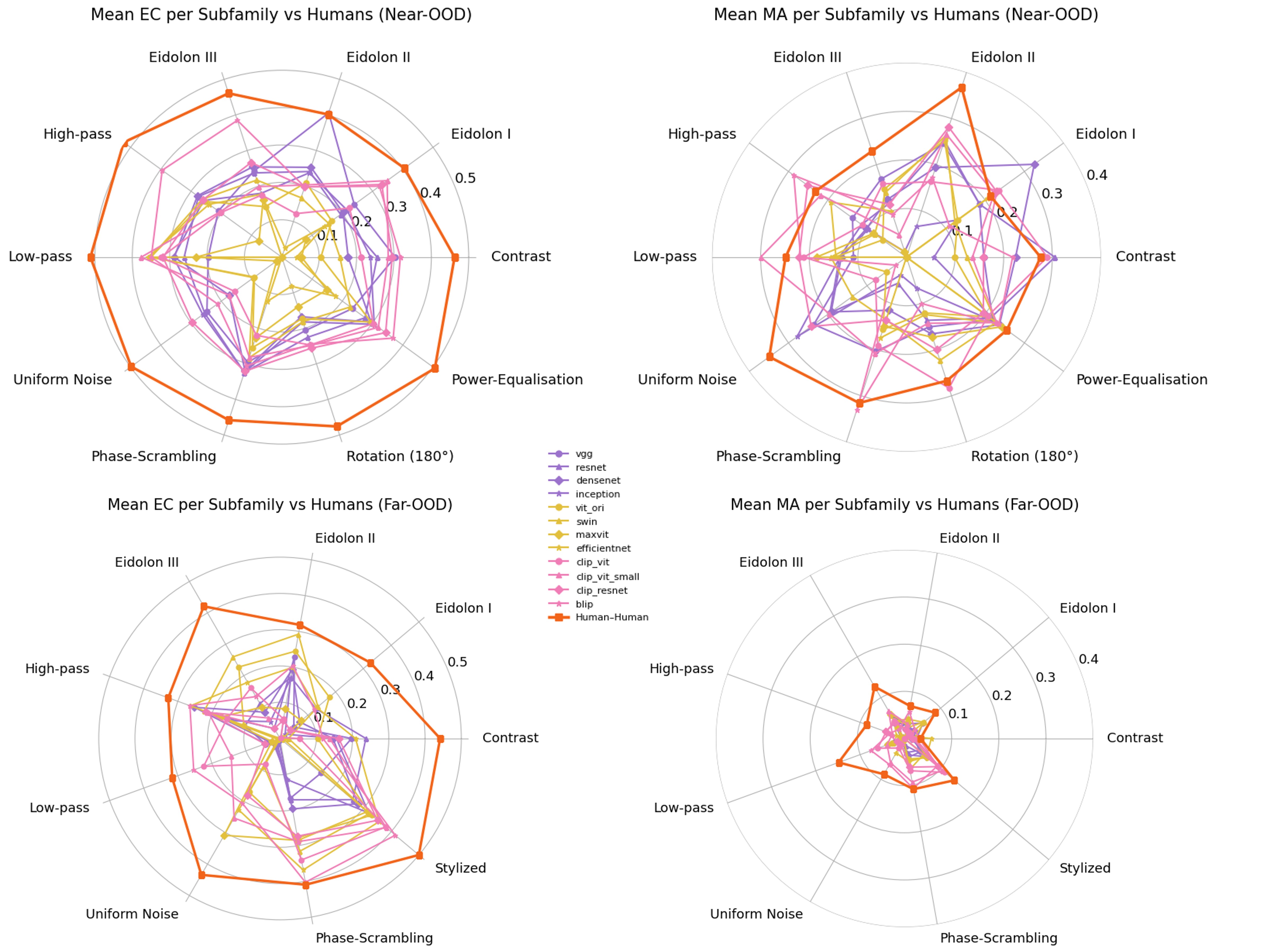}
\caption{Radar plots of mean model–human error alignment for each subfamily across distortion types, separated by OOD regime (near-OOD, EC (left top), MA (right top); far-OOD, EC (left bottom), MA (right bottom)).  Human–human alignment scores are denoted in orange.}
\label{fig:radar_subfamily}
\end{figure*}

To further examine how architectural families influence model–human alignment across distortion types, we select one representative condition for each distortion type from both near- and far-OOD regimes and visualize their alignment profiles using radar plots (Figure~\ref{fig:radar}). Similarity in the shape of the resulting curves indicates similar error patterns between architectures across distortion types. We focus here on results at the model superfamily level; results for individual subfamilies are shown in Figure~S\ref{fig:radar_subfamily}.

In the near-OOD regime, the human–human EC curve is roughly circular with relatively narrow error bars, indicating similar behavioral agreement across distortion types.
In general, VLMs and CNNs present higher EC than ViTs. 
ViTs only surpass CNNs under \textit{Low-Pass} distortions.
For MA, human–human alignment is less stable, showing wider error bars and greater variability across distortion types.
Several distortion types yield similar EC but markedly different MA patterns. For instance, \textit{Uniform Noise} and \textit{Low-Pass} produce comparable and high human-human EC, but the former yields much higher MA. 
Conversely, \textit{EidolonII} produces comparatively low EC but exceptionally high MA for both human–human and model-human pairs.
CNNs achieve both high EC and MA under \textit{Uniform Noise}, where ViTs do not. VLMs display particularly strong human alignment (especially MA) under \textit{High-Pass} and \textit{Low-Pass}, albeit with substantial variability. 
All three superfamilies show similarly strong MA under Power-Equalisation. 

In the far-OOD regime, human–human EC becomes less uniform across distortion types. 
The EC for \textit{Contrast} and \textit{Uniform Noise} changes only slightly from near- to far-OOD. 
In contrast, \textit{High-Pass} and \textit{Low-Pass} produce the largest declines in human–human EC. 
Among model families, CNNs generally exhibit the lowest EC across distortions, except under \textit{Contrast}, where they remain competitive. Under \textit{Low-Pass} and \textit{Uniform Noise}, CNNs' EC values approach zero. ViTs show a distinct advantage in the three \textit{Eidolon} conditions, whereas VLMs achieve exceptionally high EC under \textit{Low-Pass}.
For MA, human–human alignment decreases considerably compared with the near-OOD regime, suggesting that under these highly impaired conditions, it becomes much harder for humans to exhibit shared misclassifications. Nonetheless, \textit{Low-Pass}, \textit{Stylized}, and \textit{EidolonIII} maintain comparatively high human–human MA, while \textit{Contrast} leads to minimal agreement on incorrect labels. 
Across models, VLMs achieve particularly high MA under \textit{Phase-Scrambling} and \textit{Stylized}. ViTs still attain MA values close to those of VLMs in several conditions, although the absolute values remain low. CNNs consistently show very low MA across all distortions.

\subsubsection{Model-human alignment rankings shifts over OOD regimes reflect different inductive biases }
\begin{figure*}[tbhp]
\centering  
\includegraphics[width=1\linewidth]{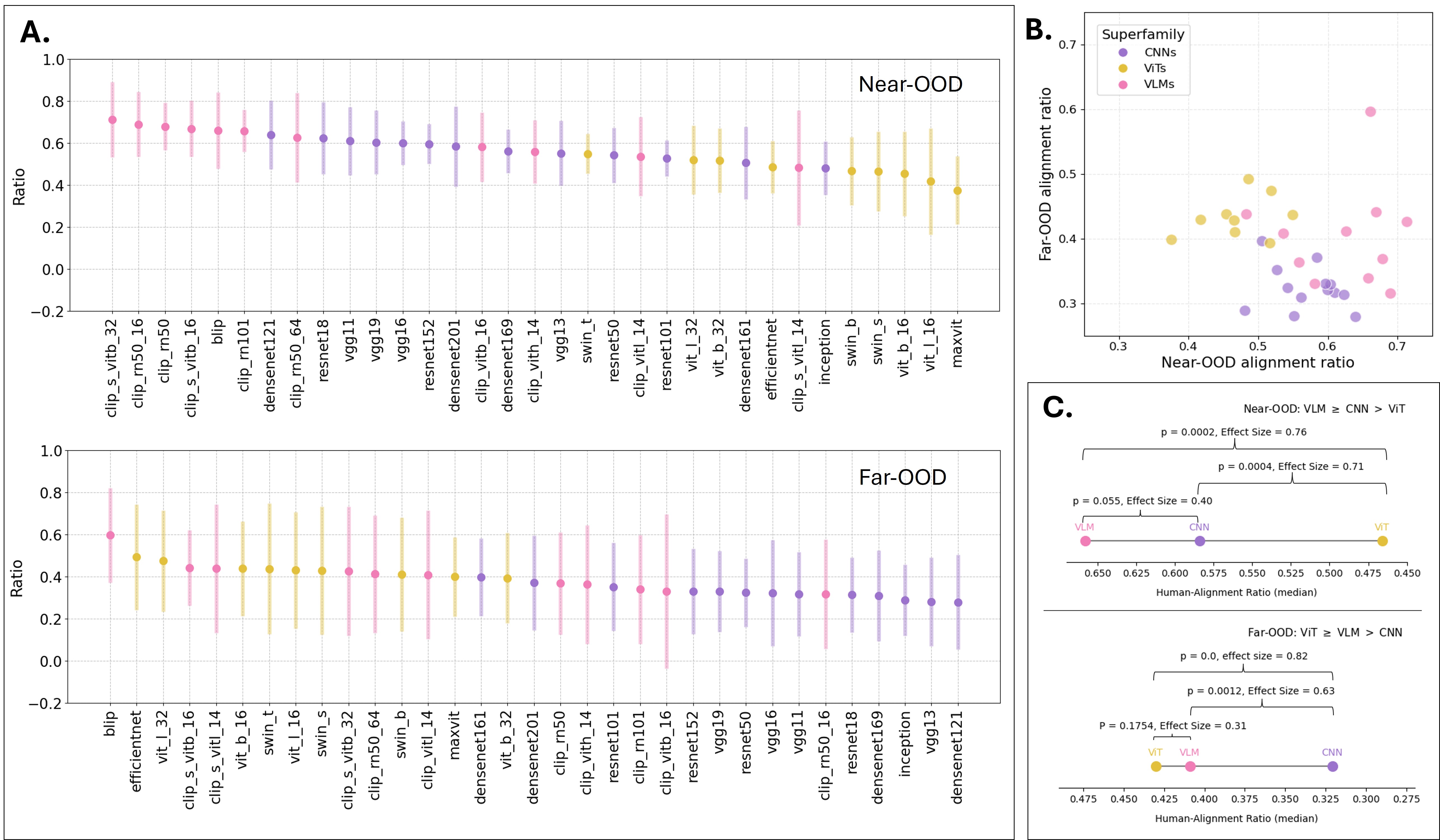}
\caption{Model–human error alignment rankings. (A) Model–human error alignment values normalised by human–human alignment ordered from most or least aligned in near- (top) and far-OOD (bottom) regimes. 
Values are the ratio of model–human alignment to human–human alignment, where higher values indicate that the error patterns between those models and humans are closer to the baseline agreement among humans.
Each dot represents a model, with vertical bars showing ±1 SD confidence interval. Colours denote model families (CNNs in purple, ViTs in yellow, VLMs in pink).
(B) Scatter plot of alignment ratio near-OOD (x-axis) and far-OOD (y-axis) conditions. (C) Pairwise statistical ranking of model superfamilies under near- and far-OOD regimes.
Each plot summarises Mann–Whitney U tests comparing the ranks between model families; “$>$” indicates a significant rank advantage ($p < 0.01$), while “$\geq$” denotes a non-significant difference. Median alignment ratios are shown only to aid visual interpretation.}
\label{fig:model_ranking}
\end{figure*}

Finally, we examine which models showed the most human-like error patterns by ranking average model-human alignment scores.
To compare how models align with humans across different model families, we normalise model–human alignment measures by human–human alignment on the corresponding condition. Specifically, for each model $m$ and condition $c$, we compute the alignment ratio 
\begin{equation}
    \rho_m^c = \frac{a_m^c}{\hat{a}^c}
\end{equation}
where $a^c_m$ is the model’s average alignment with humans (mean value of EC and MA; see Figure~S5 for EC and MA separately) and $\hat{a}^c$ is the corresponding human–human alignment score for that condition.
This ratio accounts for the inherent variability in human agreement under each distortion and evaluate models relative to a human baseline.
We denote the average of $\rho_m^c$ over all representative conditions within an OOD regime as $\bar{\rho}_m$, a summary statistic for ranking models in that regime.
The highest-ranked model in a regime is therefore the one whose alignment with humans is most comparable to the agreement observed among humans themselves.

The rankings of $\bar{\rho}_m$ for the models in near- and far-OOD regimes are shown in Figure~\ref{fig:model_ranking}A and the alignment ratios for each model in near and far-OOD regime are presented in Figure~\ref{fig:model_ranking}B.
We find considerable differences in model-human alignment between near- and far-OOD: many models that align closely with humans in near-OOD no longer rank highly in far-OOD, showing that perceptual difficulty reshapes the human-based behavioral consistency of models (Figure~\ref{fig:model_ranking}A).
In the near-OOD regime, VLMs and CNNs tend to achieve a higher alignment ratio than ViTs ($p's<0.01$), although ViTs typically outperform CNNs in terms of accuracy in the reference conditions. 
In the far-OOD regime, VLMs maintain high alignment rankings, but to a similar level to ViTs, both of which had higher human alignment over CNNs ($p<0.01$; Figure~\ref{fig:model_ranking}C).

\section{Discussion}



Model-human alignment comparisons have typically treated different distortion types and levels similarly, despite being distinct in nature. Our human-centred framework addresses this by defining an OOD spectrum that groups conditions by human perceptual difficulty, rather than by arbitrary severity levels. We show that model-human alignment depends on the type and level of distortion. VLMs are the most consistently human-aligned across both near- and far-OOD conditions, whereas CNNs are more aligned than ViTs for near OOD and ViTs are comparable to VLMs and more aligned than CNNs for far OOD.


\subsection{Why a Human-Centred OOD Spectrum}

When models are exposed to distributional shifts, the way their performance degrades can reveal their inductive biases -- the visual features they exploit for recognition. The error profile across distortion types thus serves as a behavioral signature of a system's information-processing strategies.
However, drawing valid conclusions requires a principled way to compare distortions. Distortion parameters are incommensurable across types, and prior approaches, including aggregating across severity levels, qualitatively selecting comparable distortions, or normalising to a reference model~
\cite{hendrycks2019robustness,dodge2016understanding,krizhevsky2012imagenet}, each have their limitations. We addresses this by quantifying and grouping conditions by deviation from human performance under standard, baseline performance. The resulting OOD spectrum provides a principled basis for assessing whether models fail like humans across the full spectrum of perceptual challenge. 

First, this framework allows meaningful comparisons at matched levels of human perceptual difficulty across distortion types. Without this, apparent differences in alignment may simply reflect task difficulty rather than genuine architectural biases.  For example, \textit{Eidolon I} parameter 8, \textit{High-Pass} parameter 1, and \textit{Low-Pass} parameter 5 induce similar levels of behavioral shift from baseline, even though they appear visually very different. Second, it enables the identification of appropriate conditions for evaluating model-human alignment. The method reveals four regimes: reference, representing natural variation on undistorted or lightly distorted images; near-OOD, representing moderate accuracy reduction; far-OOD, a transitional zone where performance declines at varying rates across distortion types; and extreme-OOD, where images lack sufficient information for recognition and performance falls to or below chance. Qualitatively, reference, near-OOD, and extreme-OOD emerged as strongly coherent components, while far-OOD is more diffuse, likely due to sparse sampling in this range. When both humans and models fail entirely, alignment values become uninformative. We therefore exclude extreme-OOD for evaluation. Prior work that aggregated extreme-OOD conditions may have led to conclusions that are not well supported under finer-grained OOD analysis.

\subsection{Establishing the Human–Human Behavioral Alignment Baseline}

Human-human alignment patterns across the OOD spectrum strongly suggest that near-OOD and far-OOD regimes should be evaluated separately, as systems likely rely on different strategies across these conditions.
In the near-OOD regime, humans show roughly uniform EC across distortions, suggesting that errors may primarily target ``harder" undistorted images -- those that become ambiguous under minor degradation. Such errors depend more on the stimulus than the observer. Human–human MA is also relatively high for near-OOD, meaning distortions which obscure the true class are also relatively more likely to suggest a \emph{specific} alternative, a ``false lead" that consistently draws multiple observers toward the same incorrect response, reflecting strong stimulus-driven effects.  

In the far-OOD regime, this pattern shifts. EC degrades gracefully, indicating that individual differences in responses increase as distortions become more severe, reflecting observer dependence rather than stimulus-driven effects. MA degrades more quickly than EC, suggesting that while false leads are still present, they become less common. This suggests that stronger distortions disrupt recognition in different ways across systems.
At extreme-OOD, both metrics flatline, implying no meaningful class information remains. MA in particular indicates that false leads are almost entirely absent. The absence of accurate predictions and systematic errors reflects a lack of meaningful human–human alignment where model-human alignment cannot be meaningfully assessed. Researchers should therefore avoid these conditions when drawing conclusions about model-human alignment.

These regime-specific patterns can be understood in terms of how EC and MA relate to stimulus space.
EC and MA inform us about the alignment between two systems' decision boundaries in stimulus space. They measure the degree to which two systems agree on where the distinctions between classes are, but are sensitive to where in stimulus space the concentration of stimuli sit~\cite{geirhos-2020-beyond,xu2025measuring}. 
Different distortion levels, and to a lesser degree different types, produce distinct data distributions and occupy different regions of stimulus space. 
Computing EC and MA across pooled conditions would conflate alignment patterns that may differ substantially across these regions.

CLED patterns further support the importance of perceptual difficulty, though the effect varies across regimes. In the near-OOD regime, conditions at similar difficulty levels produce similar error profiles regardless of distortion type, suggesting that perceptual difficulty dominates over the specific features impaired, although some of this may arise by chance due to the sparsity of errors when both conditions have high accuracy. In the far-OOD regime, some conditions within the same distortion type tend to cluster together, indicating that distortion-specific feature impairment may play a larger role when degradation is severe. When two conditions are both difficult and CLED is low, the relevant visual features upon which humans depend for recognition may have been impaired in similar ways.


Taken together, these findings suggest that humans employ different decision-making strategies at different levels of perceptual difficulty: in near-OOD, errors are stimulus-driven and more consistent across observers, while in far-OOD, errors become more observer-dependent and less systematic. To explore whether models align with humans in their information-processing strategies, we must therefore evaluate these regimes separately -- a model that aligns with human errors in near-OOD may not do so in far-OOD, and vice versa.

\subsection{Trajectories of Alignment: Models versus humans }
The trajectories of model accuracy and model–human alignment across distortion levels (Figure~\ref{fig:model_curves}) reveal that different distortion types elicit markedly different responses from models, both in terms of accuracy and alignment with human behavior. These divergences indicate that models exploit different visual features to humans. When a model is more sensitive than humans to a particular distortion, it suggests that the model relies more heavily on features that this distortion disrupts. Conversely, when a model is more resilient than humans, it exploits features that humans do not use or weigh less heavily.

Several patterns emerge from the trajectories. First, human accuracy is more stable across distortion types than model accuracy -- models show greater variability in how different distortions impair their performance. Second, human–human EC consistently exceeds model–human EC for almost all models, and human–human MA, in some cases, is substantially higher than model–human MA. This indicates that humans share more similar visual features for classification with each other than with models: whether one participant misclassifies a distorted image is more predictive of another participant's response than of any model's response. In short, the visual features humans rely on overlap more with each other than with those exploited by models.

\subsection{Alignment fingerprints across model families}

As noted above, humans show roughly uniform EC with other humans for all near-OOD distortions, but model–human EC is substantially lower even for these mild distortions. This indicates that models are impaired by distortions in ways that humans are not. Moreover, different model families appear to be impaired differently by various distortions, suggesting that certain families rely on visual features that others do not, or exploit features differently. Human–human MA is also relatively high for near-OOD distortions, and though lower, model–human MA remains moderately high. This implies that some false leads for humans are the same false leads for models too -- there is some overlap between the visual features exploited by machines and humans, particularly for certain distortion types such as \textit{power equalisation}. In the far-OOD regime, EC and MA are reduced more sharply for some distortion types than others, and these patterns also differ across model families.

The radar plot (Figure~\ref{fig:radar}) shows that the profile of model–human alignment across distortion types serves as a characteristic fingerprint of each model family. This fingerprint captures how a given architecture responds to different forms of degradation relative to human perception. Our quantitative analysis reinforces this interpretation. Models are more alike within superfamilies than between superfamilies, and more alike within subfamilies than with other models in the same superfamily, although whether this reflects architectural effects or if training also plays a role remains unclear. We interpret these fingerprints as evidence that model families differ in the visual features they exploit, and hence in their sensitivity to distortions that impair those features.

\subsection{Model-human alignment rankings shift across the OOD spectrum}
We find that VLMs were most consistently aligned with human behavior across the spectrum, CNNs were more aligned than ViTs for near-OOD, and ViTs were comparable to VLMs and better than CNNs for far-OOD regimes.
These regime-dependent patterns reveal how architectural biases manifest differently under varying levels of degradation. CNNs achieve high near-OOD alignment despite their texture bias~\cite{geirhos2018imagenet}, which may reflect two possibilities. First, when texture and shape cues remain partially intact in near-OOD, CNNs align with humans while relying on fundamentally different features -- one possibility is that texture and shape are often correlated in images, allowing different strategies to converge on similar outputs. Second, under moderate degradation, humans may draw on multiple features such as informative texture cues rather than relying on shape alone, thereby aligning more with CNNs.
CNNs' behavior in far-OOD catastrophically diverges from humans, with EC and MA approaching zero. This reveals that the representations CNNs develop do not align with human behavior when information is severely degraded, and that high alignment in normal to mildly degraded conditions does not guarantee alignment in general. 

ViTs uniquely demonstrate the poorest near-OOD alignment with humans despite often maintaining high or even superior accuracy compared to other architectures. 
This means that high accuracy does not necessarily translate to human-like error patterns --  ViTs can achieve strong overall performance whilst making different errors to humans.
In contrast, ViTs achieve the strongest alignment in the far-OOD regime, comparable to VLMs. 
Under severe degradation, ViTs maintain behavioral correspondence with humans compared to CNNs.
Indeed, ViTs are less texture-dependent than CNNs~\cite{tuli2021convolutional} which may contribute to their sustained far-OOD alignment: when severe distortions remove fine details, ViTs' reduced reliance on high-frequency texture features may allow them to make human-like decisions based on remaining coarse information. 

VLMs demonstrate strong alignment across both regimes.
Notably, VLMs with CNN-based visual encoders show higher alignment in near-OOD than far-OOD, reflecting architectural biases inherited from the visual backbone. BLIP's exceptional cross-regime stability may reflect its training approach: bootstrapping with synthetic captions may result in more robust semantic representations that generalize across degradation levels.
VLMs' language-based semantic knowledge may guide its behavior toward human-like decisions, especially in near-OOD. However, for far-OOD conditions, visual information may have degraded to the extent that semantic information is no longer available, leading to similar human alignment levels to the image-only version of their model.
Language-grounded representations may guide VLMs toward human-like confusion patterns. This would parallel how human semantic knowledge constrains perceptual interpretation under uncertainty~\cite{lupyan2013language,weller2019semantic}, suggesting that multimodal training provides semantic scaffolding that remains accessible even when low-level visual features become unreliable. 

Despite strong and consistent alignment in some models, human–human alignment remains a ceiling that no model approaches. 
This tells us that information processing in the human visual system and current AI models is still significantly different. This gap matters for two reasons. First, human vision exhibits graceful degradation across distortion types -- not necessarily achieving high accuracy, but maintaining consistent, slowly degrading performance rather than unpredictably excelling on some while catastrophically failing on other distortions. Models that align with human error patterns are likely exploiting features that support this stability, whereas models that diverge may rely on features that are brittle. Second, human alignment relates to trustworthiness. A model that makes human-like errors in human-like ways is more predictable and interpretable in deployment. Such failures are more acceptable because they mirror the limitations of human perception rather than reflecting opaque, non-intuitive biases. Our framework provides a tool to quantify both dimensions, measuring whether future architectures might achieve not just high accuracy, but human-like robustness and trustworthy behavior.


\section{Materials and Methods}
\subsection{Dataset}
We used the \textit{modelvshuman} benchmark~\cite{geirhos2021partial}, which provides images from 16 object categories with human behavioral data collected under controlled laboratory conditions. Images are systematically distorted across multiple severity levels to probe robustness under OOD conditions. From the 17 available distortion types, we used 14 subsets spanning diverse distortion families including uniform noise, high-pass, low-pass, contrast, phase noise, and Eidolon variants.

\subsection{Tested Models}
We evaluated 31 models spanning three architectural families: CNNs, ViTs, and VLMs.
CNNs included VGG variants (VGG-11, -13, -16, -19)~\cite{simonyan2014very}, ResNet variants (ResNet-18, -50, -101, -152)~\cite{he2016deep}, DenseNet variants (DenseNet-121, -161, -169, -201)~\cite{huang2017densely}, Inception-v3~\cite{szegedy2016rethinking}.
ViTs included four vanilla ViT variants~\cite{dosovitskiy2020image}, three Swin Transformers~\cite{liu2021swin}, MaxViT-L~\cite{tu2022maxvit} and EfficientNet~\cite{tan2019efficientnet}. All CNNs and most ViTs were pretrained on ImageNet-1K; EfficientNet was pretrained on a larger dataset.
For VLMs, we tested BLIP~\cite{li2022blip}, CLIP-ViT variants trained on DFN-2B, CLIP-ViT variants trained by OpenAI, and CLIP-ResNet variants trained by OpenAI~\cite{radford2021learning}. 
Model details are presented in the supplementary materials.\footnote{\url{https://docs.google.com/spreadsheets/d/1ldqG8LlQd_tDh9f3xdZQRA3v2UffUIhaLds2wTrhoLU/edit?usp=sharing}}

\subsection{Error Alignment Metrics}
We employed three complementary metrics to characterise model–human behavior under distribution shift. Error consistency (EC)~\cite{geirhos-2020-beyond} and misclassification agreement (MA)~\cite{xu2025measuring} were used to quantify model–human error alignment under matched visual conditions, while class-level error divergence (CLED)~\cite{xu2025measuring} was used to measure differences in error structure across OOD conditions. All metrics were computed separately for each distortion type and severity level using model predictions and human responses on the same image set.

EC measures the degree to which two systems produce correct or incorrect responses on the same stimuli. Formally, EC is defined as Cohen’s $K$ computed over binary correctness outcomes, comparing the observed agreement in correctness to the agreement expected by chance given the marginal accuracies of the two systems.
MA quantifies alignment in how errors are made when both systems misclassify the same stimuli. MA is computed by restricting the analysis to jointly misclassified instances and measuring agreement between the predicted class labels of the two systems using a multiclass Cohen’s $K$ statistic.
CLED is computed by comparing class-wise error distributions derived from confusion matrices restricted to error instances, using a weighted Jensen–Shannon divergence across classes.


\bibliographystyle{unsrt}  
\bibliography{references}

\clearpage
\section*{Appendix}

Code is available at \url{https://github.com/xubinxia/ood-spectrum}.

\subsection*{A. Statistical Evidence for the Construction of OOD Spectrum}
\begin{table}[htbp]
\centering

\begin{subtable}{0.48\linewidth}
\centering
\caption{Part I}
\label{tab:distortions1}
\begin{tabular}{l|ll}
\toprule
Distortion & raw $p$ & adjusted $p$ \\
\midrule
\textbf{contrast\_c100} & - & - \\
contrast\_c50 & 0.00731 & 0.00835 \\
contrast\_c30 & 0.00565 & 0.00680 \\
contrast\_c15 & 0.00170 & 0.00257 \\
contrast\_c10 & 0.00154 & 0.00250 \\
contrast\_c05 & 0.00154 & 0.00250 \\
contrast\_c03 & 0.00154 & 0.00250 \\
contrast\_c01 & 0.00154 & 0.00250 \\
\midrule
\textbf{rotation\_0} & - & - \\
rotation\_90 & 0.01414* & 0.01532* \\
rotation\_180 & 0.00274 & 0.00356 \\
rotation\_270 & 0.01531* & 0.01605* \\
\midrule
eidolonI\_1-10-10 & 0.00867 & 0.00971 \\
eidolonI\_2-10-10 & 0.75350* & 0.75350*\\
eidolonI\_4-10-10 & 0.01023* & 0.01127* \\
eidolonI\_8-10-10 & 0.00187 & 0.00271 \\
eidolonI\_16-10-10 & 0.00154 & 0.00250 \\
eidolonI\_32-10-10 & 0.00154 & 0.00250 \\
eidolonI\_64-10-10 & 0.00154 & 0.00250 \\
eidolonI\_128-10-10 & 0.00154 & 0.00250 \\
\midrule
eidolonII\_1-3-10 & 0.00433 & 0.00541 \\
eidolonII\_2-3-10 & 0.00170 & 0.00257 \\
eidolonII\_4-3-10 & 0.00249 & 0.00330 \\
eidolonII\_8-3-10 & 0.00154 & 0.00250 \\
eidolonII\_16-3-10 & 0.00154 & 0.00250 \\
eidolonII\_32-3-10 & 0.00154 & 0.00250 \\
eidolonII\_64-3-10 & 0.00154 & 0.00250 \\
eidolonII\_128-3-10 & 0.00154 & 0.00250 \\
\midrule
eidolonIII\_1-0-10 & 0.00732 & 0.00835 \\
eidolonIII\_2-0-10 & 0.00329 & 0.00420 \\
eidolonIII\_4-0-10 & 0.00154 & 0.00250 \\
eidolonIII\_8-0-10 & 0.00154 & 0.00250 \\
eidolonIII\_16-0-10 & 0.00154 & 0.00250 \\
eidolonIII\_32-0-10 & 0.00154 & 0.00250 \\
eidolonIII\_64-0-10 & 0.00154 & 0.00250 \\
eidolonIII\_128-0-10 & 0.00154 & 0.00250 \\
\bottomrule
\end{tabular}
\end{subtable}
\hfill
\begin{subtable}{0.48\linewidth}
\centering
\caption{Part II}
\label{tab:distortions2}
\begin{tabular}{l|ll}
\toprule
Distortion & raw $p$  & adjusted $p$ \\
\midrule
\textbf{high-pass\_{inf}} & - & -\\
high-pass\_3 & 0.00565 & 0.00680 \\
high-pass\_1.5 & 0.00154 & 0.00250 \\
high-pass\_1 & 0.00154 & 0.00250 \\
high-pass\_0.7 & 0.00154 & 0.00250 \\
high-pass\_0.55 & 0.00154 & 0.00250 \\
high-pass\_0.45 & 0.00154 & 0.00250 \\
high-pass\_0.4 & 0.00154 & 0.00250 \\
\midrule
\textbf{low-pass\_0} & - & - \\
low-pass\_1 & 0.13754* & 0.13972* \\
low-pass\_3 & 0.00249 & 0.00330 \\
low-pass\_5 & 0.00154 & 0.00250 \\
low-pass\_7 & 0.00154 & 0.00250 \\
low-pass\_10 & 0.00154 & 0.00250 \\
low-pass\_15 & 0.00154 & 0.00250 \\
low-pass\_40 & 0.00154 & 0.00250 \\
\midrule
uniform-noise\_0.00 & 0.00206 & 0.00291 \\
uniform-noise\_0.03 & 0.00730 & 0.00835 \\
uniform-noise\_0.05 & 0.00170 & 0.00257 \\
uniform-noise\_0.10 & 0.00187 & 0.00271 \\
uniform-noise\_0.20 & 0.00154 & 0.00250 \\
uniform-noise\_0.35 & 0.00154 & 0.00250 \\
uniform-noise\_0.60 & 0.00154 & 0.00250 \\
uniform-noise\_0.90 & 0.00154 & 0.00250 \\
\midrule
\textbf{phase-scrambling\_0} & - & - \\
phase-scrambling\_30 & 0.13757* & 0.13972* \\
phase-scrambling\_60 & 0.00226 & 0.00313 \\
phase-scrambling\_90 & 0.00154 & 0.00250 \\
phase-scrambling\_120 & 0.00154 & 0.00250 \\
phase-scrambling\_150 & 0.00154 & 0.00250 \\
phase-scrambling\_180 & 0.00154 & 0.00250 \\
\midrule
\textbf{power-equalisation\_0} &- & - \\
power-equalisation\_pow & 0.00154 & 0.00250 \\
\midrule
\textbf{colour\_bw} & - & -\\
\midrule
sketch & 0.01486* & 0.01583* \\
\midrule
stylized & 0.00048 & 0.00250 \\
\bottomrule
\end{tabular}
\end{subtable}
\vspace{5mm}
\caption{Mann--Whitney U test $p$-values (raw and BH-corrected) for all distortion conditions. 
Each $p$-value indicates whether the distribution of human accuracies under the distortion differs significantly from the non-distorted baseline. 
Lower values (e.g., $p < 0.01$) suggest a statistically significant shift in prediction behavior. The adjusted $p$ values (corrected by the Benjamini–Hochberg procedure) are used to reject conditions under false discovery rate control at $\alpha = 0.01$. 
Bold rows denote conditions with no distortion applied (reference condition/baseline). Asterisks (*) mark conditions with raw $p > 0.01$ (all of them also rejected the null hypothesis after the BH-procedure).}
\label{tab:mw_test}
\end{table}

\begin{table}[htbp]
\centering

\begin{subtable}{0.48\linewidth}
\centering
\caption{Part I}
\label{tab:binomial1}
\begin{tabular}{l|ll}
\toprule
Distortion & adjusted $p$ & reject $H_0$ \\
\midrule
contrast\_c50 & 0 & True \\
contrast\_c30 & 0 & True \\
contrast\_c15 & 0 & True \\
contrast\_c10 & 4.6e-283 & True \\
contrast\_c05 & 3.9e-106 & True \\
contrast\_c03 & 1.5e-15 & True \\
contrast\_c01 & 0.47 & \textbf{False} \\
\midrule
rotation\_90 & 0 & True \\
rotation\_180 & 0 & True \\
rotation\_270 & 0 & True \\
\midrule
eidolonI\_1-10-10 & 0 & True \\
eidolonI\_2-10-10 & 0 & True \\
eidolonI\_4-10-10 & 0 & True \\
eidolonI\_8-10-10 & 0 & True \\
eidolonI\_16-10-10 & 6.4e-162 & True \\
eidolonI\_32-10-10 & 2.4e-40 & True \\
eidolonI\_64-10-10 & 5.6e-09 & True \\
eidolonI\_128-10-10 & 0.074 & \textbf{False} \\
\midrule
eidolonII\_1-3-10 & 0 & True \\
eidolonII\_2-3-10 & 0 & True \\
eidolonII\_4-3-10 & 0 & True \\
eidolonII\_8-3-10 & 8.6e-253 & True \\
eidolonII\_16-3-10 & 1.4e-43 & True \\
eidolonII\_32-3-10 & 0.04 & \textbf{False} \\
eidolonII\_64-3-10 & 0.2 & \textbf{False} \\
eidolonII\_128-3-10 & 0.24 & \textbf{False} \\
\midrule
eidolonIII\_1-0-10 & 0 & True \\
eidolonIII\_2-0-10 & 0 & True \\
eidolonIII\_4-0-10 & 0 & True \\
eidolonIII\_8-0-10 & 6.2e-139 & True \\
eidolonIII\_16-0-10 & 2.1e-16 & True \\
eidolonIII\_32-0-10 & 2.2e-05 & True \\
eidolonIII\_64-0-10 & 0.2 & \textbf{False} \\
eidolonIII\_128-0-10 & 0.52 & \textbf{False} \\
\bottomrule
\end{tabular}
\end{subtable}
\hfill
\begin{subtable}{0.48\linewidth}
\centering
\caption{Part II}
\label{tab:binomial2}
\begin{tabular}{l|ll}
\toprule
Distortion & adjusted $p$ & reject $H_0$ \\
\midrule
high-pass\_3 & 0 & True \\
high-pass\_1.5 & 0 & True \\
high-pass\_1 & 1.9e-299 & True \\
high-pass\_0.7 & 3.9e-106 & True \\
high-pass\_0.55 & 4.1e-24 & True \\
high-pass\_0.45 & 0.13 & \textbf{False} \\
high-pass\_0.4 & 0.41 & \textbf{False} \\
\midrule
low-pass\_1 & 0 & True \\
low-pass\_3 & 0 & True \\
low-pass\_5 & 5.4e-312 & True \\
low-pass\_7 & 5e-275 & True \\
low-pass\_10 & 4.3e-123 & True \\
low-pass\_15 & 8.1e-32 & True \\
low-pass\_40 & 0.52 & \textbf{False} \\
\midrule
uniform-noise\_0.00 & 0 & True \\
uniform-noise\_0.03 & 0 & True \\
uniform-noise\_0.05 & 0 & True \\
uniform-noise\_0.10 & 0 & True \\
uniform-noise\_0.20 & 0 & True \\
uniform-noise\_0.35 & 6.6e-134 & True \\
uniform-noise\_0.60 & 4.1e-15 & True \\
uniform-noise\_0.90 & 0.098 & \textbf{False} \\
\midrule
phase-scrambling\_30 & 0 & True \\
phase-scrambling\_60 & 0 & True \\
phase-scrambling\_90 & 0 & True \\
phase-scrambling\_120 & 6e-129 & True \\
phase-scrambling\_150 & 6.4e-06 & True \\
phase-scrambling\_180 & 6.9e-05 & True \\
\midrule
power-equalisation\_pow & 0 & True \\
sketch & 0 & True \\
\midrule
stylized & 0 & True \\
\bottomrule
\end{tabular}
\end{subtable}

\vspace{5mm}
\caption{Results of binomial tests under the null hypothesis $P(1/16)$.  }
\label{tab:binomial_test}
\end{table}

\begin{figure}[htbp]
    \centering
    \begin{subfigure}{0.46\textwidth}
        \centering\includegraphics[width=0.9\linewidth]{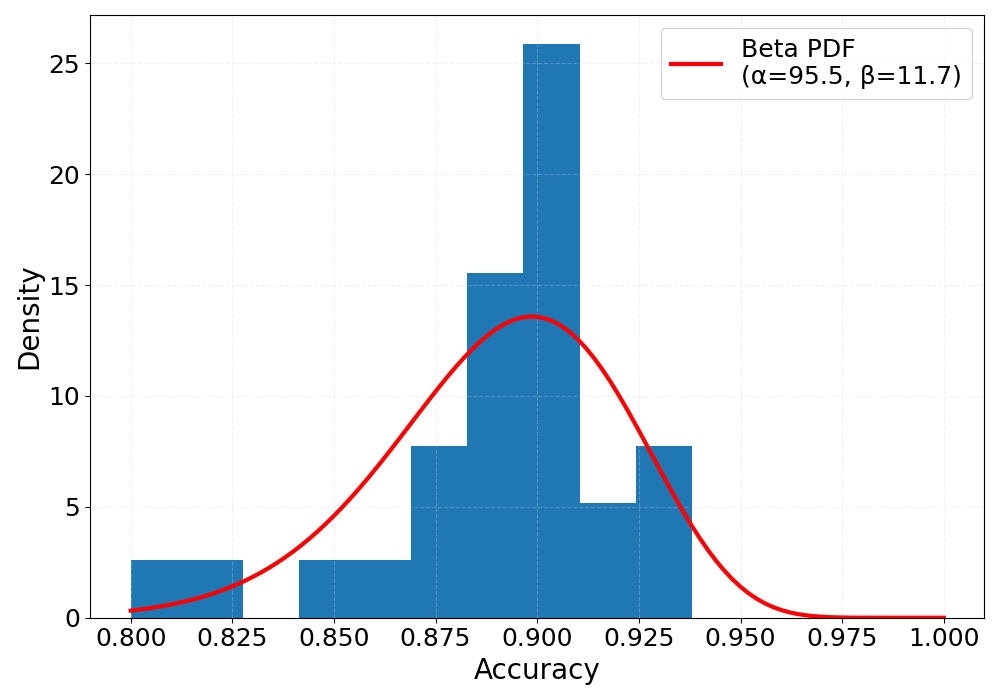}
        \caption{Histogram of raw accuracy values with fitted Beta distribution.}
        \label{fig:acc_dist}
    \end{subfigure}
    \hfill
    \begin{subfigure}{0.46\textwidth}
        \centering
        \includegraphics[width=0.9\linewidth]{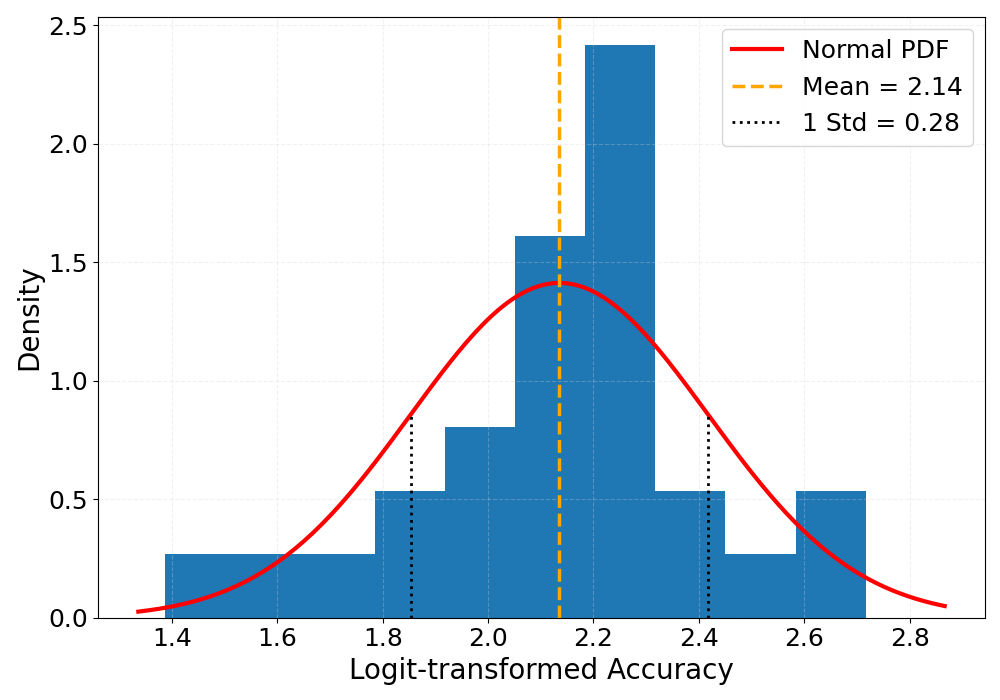} 
        \caption{Histogram of logit-transformed accuracies with fitted normal distribution.}
        \label{fig:acc_logit_dist}
    \end{subfigure}
    \caption{ Histogram of accuracy on the uncorrupted datasets before and after logit transformation. \textbf{(a)} Accuracy values lie in the bounded interval [0, 1], and those values for the uncorrupted set are usually very high, leading to a left-skewed distribution near 1. The fitted Beta distribution illustrates this skew. \textbf{(b)} After applying the logit transformation, the distribution becomes more symmetric, resembling a normal distribution. This transformation allows standardised comparisons using parametric methods.}
    \label{fig:acc_logit_dist_combined}
\end{figure}

To assess whether human accuracy scores were normally distributed, we first plot the probability density of the 28 accuracy values in Figure~\ref{fig:acc_dist}. As shown, the accuracy scores are very close to 1. Due to this upper bound, the distribution is left-skewed, which is not ideal for parametric tests that assume normality and symmetric variance. To deal with this issue, we apply the logit transformation:
\begin{equation}
    logit(a) = ln(\frac{a}{1-a}).
\end{equation}
This maps the interval $[0, 1]$ onto the entire real number space $(-\infty, +\infty)$. For simplification, we denote the $logit(a)$ as $l_a$ in the following. The histogram for $l_a$ is illustrated in Figure~\ref{fig:acc_logit_dist}, which resembles a normal distribution.

To support this statistically, we conducted a set of normality tests (Table~\ref{tab:normality_tests}) and show that for the raw accuracies, the null hypothesis should be rejected, while for the logits, we cannot reject the null hypothesis. This supports the use of effect size for the standardised comparisons that follow.

\begin{table}[htbp]
    \centering

    \begin{tabular}{lccc}
        \toprule
        & Shapiro--Wilk & D'Agostino--Pearson & Lilliefors \\
        \midrule
        Accuracy ($a$)             
        & 0.004 & 0.001 & 0.021 \\
        Logit-transformed Accuracy ($l_a$)     & 0.194 & 0.135 & 0.192 \\
        \bottomrule
    \end{tabular}
        \vspace{0.5em}

       \caption{P-values from three normality tests (Shapiro--Wilk, D'Agostino--Pearson, and Lilliefors) applied to both raw accuracy values and logit-transformed accuracy. While the raw accuracy values significantly deviate from normality ($p < 0.05$), the logit-transformed accuracies yield p-values above the 0.05 threshold across all tests. This indicates that we cannot reject the null hypothesis that $l_a$ is drawn from a normal distribution.}
        \label{tab:normality_tests}
\end{table}

\subsection*{B. Formal Definitions of Error Alignment Metrics}
This section provides complete definitions for the three behavioral alignment metrics used to characterise error patterns between classification systems.

Consider a dataset $\mathcal{D} = \{(x_n, t_n)\}_{n=1}^{N}$ comprising inputs $x_n \in \mathcal{X}$ and target labels $t_n \in \mathcal{Y}$, where $\mathcal{Y}$ contains $C$ classes. Let $A$ and $B$ denote two classification systems that produce predictions $y_n^A$ and $y_n^B$ respectively for each input $x_n$. Let $p_A$ and $p_B$ denote the accuracies of systems $A$ and $B$ on $\mathcal{D}$.

\subsubsection*{B.1 Error Consistency (EC)}
Error Consistency~\cite{geirhos-2020-beyond} measures the degree to which two systems produce correct or incorrect responses on the same stimuli, beyond what would be expected by chance given their individual accuracies.

First, we need to define  the following instance counts on the dataset $\mathcal{D}$:
\begin{itemize}
    \item  $N_c$: number of **jointly correct** instances where $t_n = y_n^A = y_n^B$
\item $N_e$: number of \textit{jointly incorrect} instances where $y_n^A \neq t_n$ and $y_n^B \neq t_n$
\end{itemize}

The \textit{observed agreement rate} is the proportion of instances on which both systems agree in their correctness:
\begin{equation}
    p_{\text{obs}} = \frac{N_c + N_e}{N}
\end{equation}

The \textit{expected agreement rate by chance}, assuming independence between the two systems' correctness, is:
\begin{equation}
  p_{\text{exp}} = p_A \cdot p_B + (1 - p_A)(1 - p_B)  
\end{equation}

where $p_A$ and $p_B$ are the accuracies of systems $A$ and $B$ respectively.

Error Consistency is then defined as Cohen's kappa ($\kappa$) computed over these values:
\begin{equation}
   \text{EC}(A, B) = \frac{p_{\text{obs}} - p_{\text{exp}}}{1 - p_{\text{exp}}} 
\end{equation}

\subsubsection*{B.2 Misclassification Agreement (MA)}

Misclassification Agreement~\cite{xu2025measuring} quantifies alignment in the specific errors made when both systems misclassify the same stimuli, addressing the limitation that EC treats all joint errors identically regardless of the predicted classes.

The \textit{error set} for a system $g \in \{A, B\}$ is defined as as:

$$\mathcal{D}_g^{\text{err}} = \{(x_n, t_n) \in \mathcal{D} : y_n^g \neq t_n\}$$

The \textit{joint error set} is the intersection:

$$\mathcal{D}_{A,B}^{\text{err}} = \mathcal{D}_A^{\text{err}} \cap \mathcal{D}_B^{\text{err}}$$

We then define the \textit{error agreement matrix} $M^{\text{err}} \in \mathbb{Z}_+^{C \times C}$ where the $(i,j)$-th element counts joint error instances with predictions $y_n^A = i$ and $y_n^B = j$:

$$[M^{\text{err}}]_{ij} = \left| \{(x_n, t_n) \in \mathcal{D}_{A,B}^{\text{err}} : y_n^A = i, \, y_n^B = j\} \right|$$

Let $N^{\text{err}} = |\mathcal{D}_{A,B}^{\text{err}}|$ denote the total number of joint errors.

The \textit{observed error-agreement rate} is the proportion of joint errors on which $A$ and $B$ predict the same (incorrect) class:
\begin{equation}
    \tilde{p}_o = \frac{N_O^{\text{err}}}{N^{\text{err}}} = \frac{\sum_{i=1}^{C} [M^{\text{err}}]_{ii}}{N^{\text{err}}}
\end{equation}

where $N_O^{\text{err}}$ is the number of joint errors on which both systems agree (i.e., the trace of $M^{\text{err}}$).

The \textit{expected error-agreement rate by chance}, under the null hypothesis that predictions are independent on the joint error set, is:
\begin{equation}
  \tilde{p}_e = \sum_{i=1}^{C} \hat{p}_i^{(A)} \cdot \hat{p}_i^{(B)}  
\end{equation}

where $\hat{p}_i^{(g)}$ is the estimated probability that system $g$ predicts class $i$ on $\mathcal{D}_{A,B}^{\text{err}}$:
\begin{equation}
    \hat{p}_i^{(A)} = \frac{\sum_{j=1}^{C} [M^{\text{err}}]_{ij}}{N^{\text{err}}}, \quad \hat{p}_i^{(B)} = \frac{\sum_{j=1}^{C} [M^{\text{err}}]_{ji}}{N^{\text{err}}}
\end{equation}
(i.e., row and column marginals of $M^{\text{err}}$ respectively).

Then, Misclassification Agreement is the multiclass Cohen's kappa of the error agreement matrix:
\begin{equation}
   \text{MA}(A, B) = \kappa(M^{\text{err}}) = \frac{\tilde{p}_o - \tilde{p}_e}{1 - \tilde{p}_e} 
\end{equation}

\subsubsection*{B.3 Class-Level Error Divergence (CLED)}

Class-Level Error Divergence~\cite{xu2025measuring} measures the divergence of the distribution between the error patterns of two systems at the class level, without requiring instance-level correspondence. This enables comparison even when only aggregate confusion data is available.

For system $g \in \{A, B\}$, define the \textit{error confusion matrix} $F_g^{\text{err}} \in \mathbb{Z}_+^{C \times C}$ where the $(i,j)$-th element counts error instances with true class $i$ and predicted class $j$:

$$[F_g^{\text{err}}]_{ij} = \left| \{(x_n, t_n) \in \mathcal{D}_g^{\text{err}} : t_n = i, \, y_n^g = j\} \right|$$

Note that diagonal elements $[F_g^{\text{err}}]_{ii} = 0$ by construction (correct predictions are excluded).

For each true class $i$, collect the row of counts into vector $\mathbf{f}_i^g = ([F_g^{\text{err}}]_{i1}, \ldots, [F_g^{\text{err}}]_{iC})^\top$. The \textit{estimated error distribution} for system $g$ on class $i$ is computed as the expectation of a posterior Dirichlet distribution with prior $\boldsymbol{\alpha} \in \mathbb{R}_+^C$:
\begin{equation}
    \hat{\boldsymbol{\pi}}_i^g = \frac{\mathbf{f}_i^g + \boldsymbol{\alpha}}{\mathbf{1}^\top (\mathbf{f}_i^g + \boldsymbol{\alpha})}
\end{equation}

where $\mathbf{1}$ is the vector of all ones. A symmetric prior $\boldsymbol{\alpha} = 0.5 \cdot \mathbf{1}$ (Jeffreys prior) is typically used.

The Jensen-Shannon Divergence (JSD) between two probability distributions $P$ and $Q$ is:
\begin{equation}
    \text{JSD}(P, Q) = \frac{1}{2} D_{\text{KL}}(P \| M) + \frac{1}{2} D_{\text{KL}}(Q \| M)
\end{equation}

where $M = \frac{1}{2}(P + Q)$ is the mixture distribution and $D_{\text{KL}}$ denotes the Kullback-Leibler divergence:
\begin{equation}
   D_{\text{KL}}(P \| Q) = \sum_i P(i) \log \frac{P(i)}{Q(i)} 
\end{equation}

JSD is symmetric, bounded in $[0, 1]$ (when using $\log_2$), and well-defined even when distributions have zero entries.

The \textit{Class-Level Error Divergence} aggregates JSD across all classes with sample-size weighting:
\begin{equation}
    \text{CLED}(A, B) = \sum_{i=1}^{C} w_i \cdot \text{JSD}(\hat{\boldsymbol{\pi}}_i^A, \hat{\boldsymbol{\pi}}_i^B)
\end{equation}

where the weight $w_i$ for class $i$ is the proportion of total errors (across both systems) with ground-truth class $c$, normalised by the total number of errors:
\begin{equation}
    w_i = \frac{n_i^A + n_i^B}{\sum_{j=1}^{I} (n_j^A + n_j^B)}
\end{equation}

\subsection*{C. Patterns for Error Alignment}

\begin{figure}[htbp]
    \centering
    \begin{subfigure}{0.49\textwidth}
        \centering\includegraphics[width=1\linewidth]{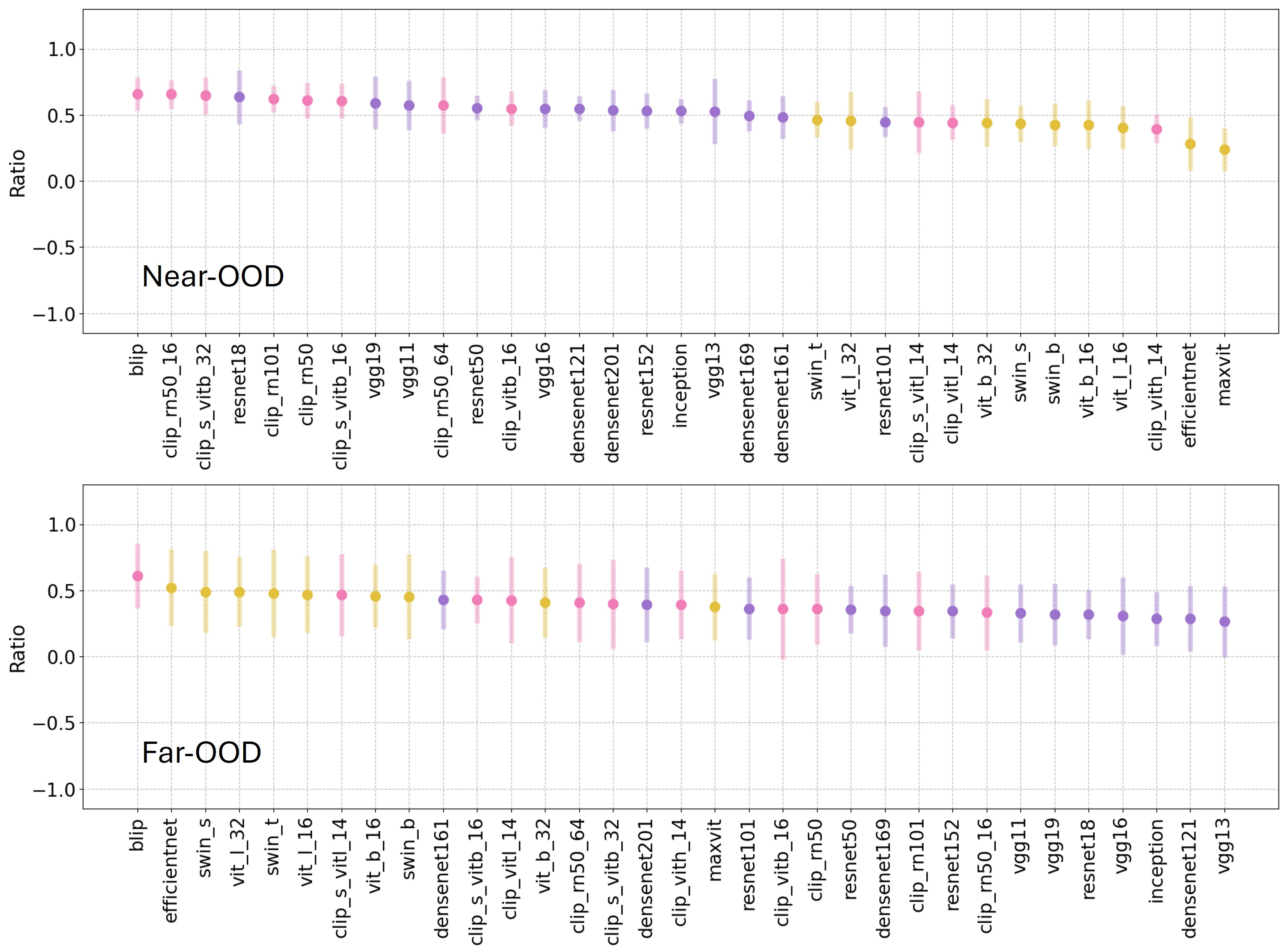}
        \caption{Error Consistency (EC)}
    \end{subfigure}
    \hfill
    \begin{subfigure}{0.49\textwidth}
        \centering
        \includegraphics[width=1\linewidth]{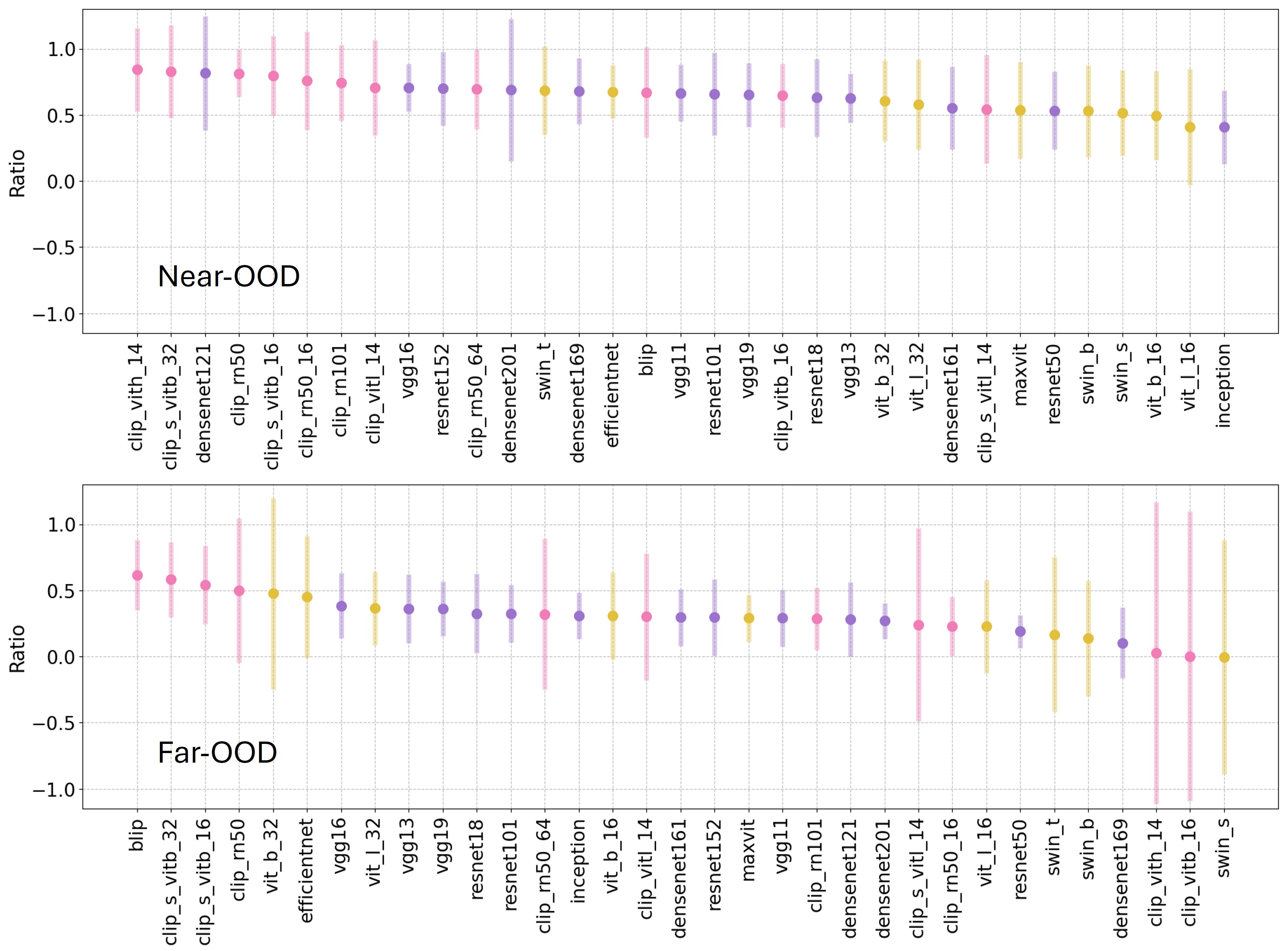} 
        \caption{Misclassification Agreement (MA)}
    \end{subfigure}
    \caption{The ranking for each metric. }
    \label{fig:ranking_each_metric}
\end{figure}

\end{document}